\documentclass[10pt,twocolumn,letterpaper]{article}

\usepackage[pagenumbers]{cvpr} 

\usepackage{times}
\usepackage{epsfig}
\usepackage{graphicx}
\usepackage{amsmath}
\usepackage{amssymb}
\usepackage{booktabs}
\usepackage{multirow}
\usepackage{makecell}
\usepackage{enumitem}
\usepackage{wrapfig}
\usepackage{subcaption}
\usepackage{mathtools}
\usepackage{bbm}
\usepackage{microtype}
\usepackage[dvipsnames]{xcolor}

\setlist[itemize]{leftmargin=*,topsep=2pt,itemsep=2pt}
\setlist[enumerate]{leftmargin=*,topsep=2pt,itemsep=2pt}

\graphicspath{{figs/}}

\usepackage[misc]{ifsym}

\definecolor{cvprblue}{rgb}{0.21,0.49,0.74}
\usepackage[pagebackref,breaklinks,colorlinks,allcolors=cvprblue]{hyperref}
\usepackage{graphicx}
\usepackage{float}
\usepackage{overpic}
\usepackage{wrapfig}
\usepackage{cuted}
\usepackage{graphicx} 
\usepackage{caption}
\usepackage{booktabs}
\usepackage{siunitx}
\usepackage{xcolor}
\usepackage{multirow}
\usepackage{colortbl}

\definecolor{gray60}{gray}{.60}
\definecolor{gray92}{gray}{.92}
\definecolor{gray94}{gray}{.94}
\definecolor{gray96}{gray}{.96}

\definecolor{lightblue}{RGB}{224,239,251}

\usepackage[utf8]{inputenc}
\def\paperID{44929}
\def\confName{CVPR}
\def\confYear{2026}

\title{CARE-Edit: Condition-Aware Routing of Experts for Contextual Image Editing}

\author{
\begin{tabular}[t]{@{}c@{}}
Yucheng Wang\thanks{Equal contribution. \quad \textrm{\Letter} Corresponding author.} \quad Zedong Wang\footnotemark[1] \quad Yuetong Wu  \quad Yue Ma \quad Dan Xu${^{\textrm{\Letter}}}$
\end{tabular}\\[0.7ex]
\begin{tabular}[t]{@{}c@{}}
The Hong Kong University of Science and Technology
\end{tabular}\\[0.2ex]
 {\tt\small
 \{ywangls, zwangmw, ywufe\}@connect.ust.hk \quad
 mayuefighting@gmail.com \quad
 danxu@cse.ust.hk
 }}

\begin{document}
\maketitle

\begin{strip} 
    \centering
    \vspace{-50pt}
    \captionsetup{type=figure}
    \includegraphics[width=\textwidth]{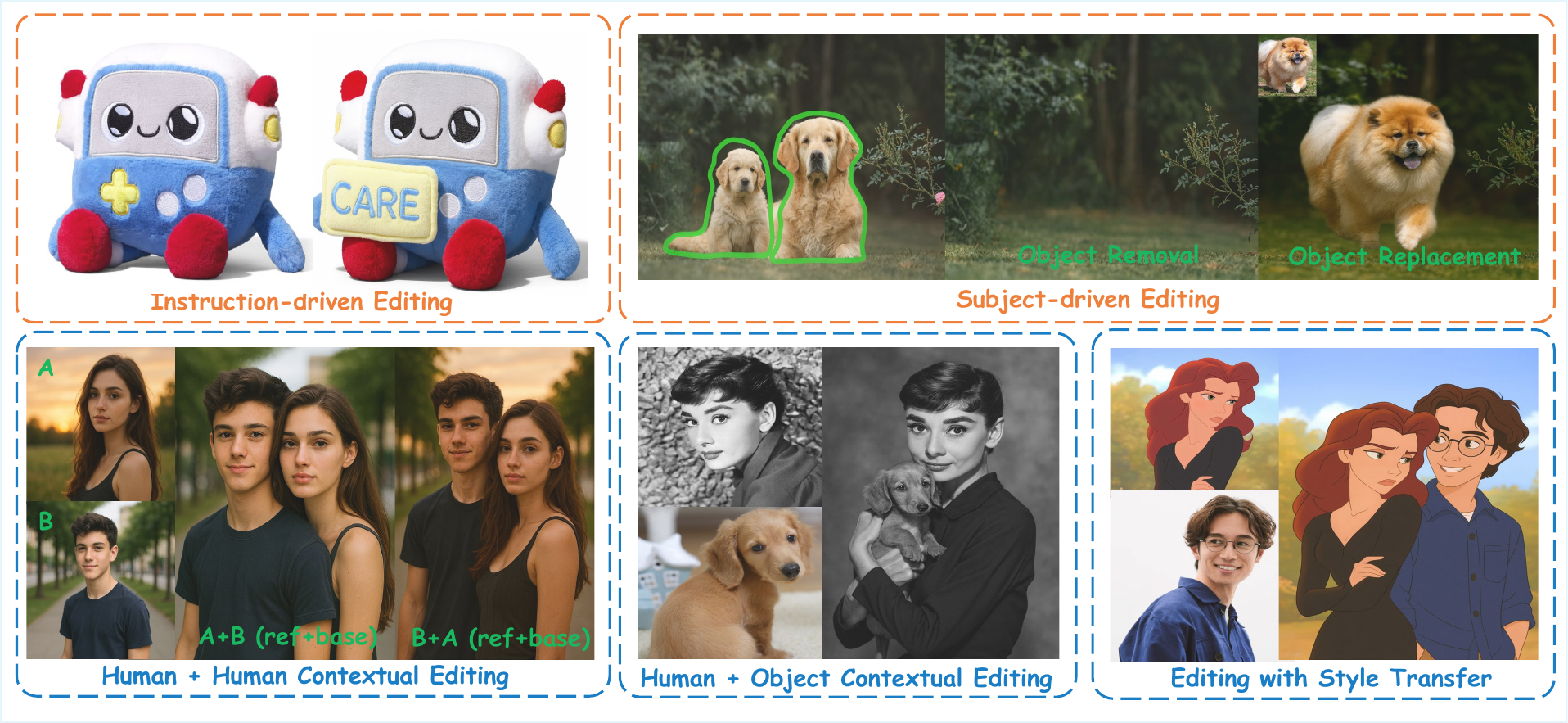} 
    \vspace{-2.20em}
    \caption{CARE-Edit routes diffusion tokens to heterogeneous experts conditioned on text, mask, and reference signals, enabling high-fidelity editing across diverse tasks: instruction-driven modification, subject removal and replacement, scene reconfiguration, and style transfer.}
    \label{fig:teaser}
\end{strip}

\begin{abstract}
Unified diffusion editors often rely on a fixed, shared backbone for diverse tasks, suffering from task interference and poor adaptation to heterogeneous demands (e.g., local vs global, semantic vs photometric). In particular, prevalent ControlNet and OmniControl variants combine multiple conditioning signals (e.g., text, mask, reference) via static concatenation or additive adapters which cannot dynamically prioritize or suppress conflicting modalities, thus resulting in artifacts like color bleeding across mask boundaries, identity or style drift, and unpredictable behavior under multi-condition inputs. To address this, we propose \textbf{\underline{C}}ondition-\textbf{\underline{A}}ware \textbf{\underline{R}}outing of \textbf{\underline{E}}xperts (\textbf{CARE}-Edit) that aligns model computation with specific editing competencies. At its core, a lightweight latent-attention router assigns encoded diffusion tokens to four specialized experts--Text, Mask, Reference, and Base--based on multi-modal conditions and diffusion timesteps: (i) a Mask Repaint module first refines coarse user-defined masks for precise spatial guidance; (ii) the router applies sparse top-K selection to dynamically allocate computation to the most relevant experts; (iii) a Latent Mixture module subsequently fuses expert outputs, coherently integrating semantic, spatial, and stylistic information to the base images. Experiments validate CARE-Edit's strong performance on contextual editing tasks, including erasure, replacement, text-driven edits, and style transfer. Empirical analysis further reveals task-specific behavior of specialized experts, showcasing the importance of dynamic, condition-aware processing to mitigate multi-condition conflicts. The project page and source code are available at \href{https://care-edit.github.io/}{LINK}.

\end{abstract}
\section{Introduction}

Diffusion-based models have radically transformed image editing, making diverse tasks like localized object replacement, global style adjustment, and text-driven content insertion more accessible and with higher quality~\citep{ho2020ddpm, song2020score, rombach2022latent, saharia2022photorealistictexttoimagediffusionmodels}.
However, most unified editors process all edits with a fixed, shared backbone, which struggles to adapt to heterogeneous demands in practice. This could potentially result in artifacts such as color bleeding at mask boundaries, identity or style drift when using reference, or inconsistent behavior when multiple, conflicting conditions must be satisfied~\cite{tumanyan2022plugandplaydiffusionfeaturestextdriven,wang2024instantidzeroshotidentitypreservinggeneration}.

One of the key reasons for this limitation lies in the static fusion of control signals. Popular methods like ControlNet and OmniControl-style variants integrate multi-modal inputs (\eg, text prompts, masks, reference images, or sketches) via simple concatenation or additive adapters attached to the backbone~\citep{zhang2023controlnet, zhang2024omnicontrol, mou2023t2iadapterlearningadaptersdig}. While effective for single-condition guidance, it is not \textit{condition-aware}, \ie, it cannot allocate the finite model capacity \textit{adaptively} to multiple, heterogeneous input conditions. As such, competing signals might not be properly resolved, inducing issues appeared in Figure~\ref{fig:maincompare}: text semantics may override mask constraints, reference identity or style may be misapplied, and global adjustments can spill into regions that should be preserved. More importantly, the importance of these signals changes over the diffusion trajectory, from semantic layout in early steps to boundary refinement and style consistency in later ones~\citep{ho2020ddpm, nichol2021improved, ho2022classifierfree}, yet static methods offer little mechanism to adapt this balance.

In this paper, we present CARE-Edit, a \textbf{\underline{C}}ondition-\textbf{\underline{A}}ware \textbf{\underline{R}}outing of \textbf{\underline{E}}xperts framework to tackle above challenges. Instead of forcing all signals through one shared pathway, a lightweight latent-attention router conditions on the prompt, mask statistics, reference features, and diffusion timestep to \emph{dynamically} dispatch tokens to four specialized experts: (i) \emph{Text} for semantic reasoning and synthesis, (ii) \emph{Mask} for spatial precision and boundary refinement, (iii) \emph{Reference} for identity/style transfer, and (iv) \emph{Base} for global coherence and background preservation. Sparse top-$K$ routing enables token-wise and timestep-aware prioritization, while a persistent shared expert stabilizes training and prevents routing collapse, following practices in previous studies~\citep{lepikhin2020gshard, roller2021hash, fedus2022switch}.

\begin{figure*}[t]
    \centering    \includegraphics[width=\textwidth]{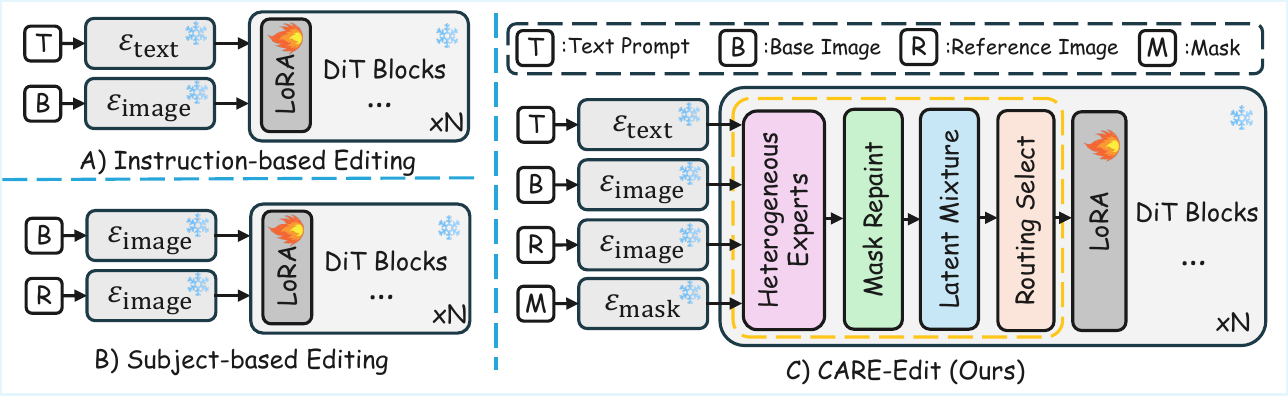}
    \vspace{-1.40em}
    \caption{Overview of contextual image editing paradigm. (A)~Instruction‑based editing guides modifications via a text prompt $\mathbf{T}$ and a base image $\mathbf{B}$. (B)~Subject‑based editing uses the base $\mathbf{B}$ and a references image $\mathbf{R}$ to preserve identity or style. 
    (C)~CARE‑Edit incorporates all these modalities $(\mathbf{T}, \mathbf{B}, \mathbf{R})$ and the user-defined mask $\mathbf{M}$ in a diffusion transformer (DiT) backbone with condition‑aware routing of experts.}
    \label{fig:overview}
    \vspace{-0.60em}
\end{figure*}

To further push its limits, we introduce two complementary designs as in Figure~\ref{fig:overview}. First, we facilitate interactions via Mask Repaint (Sec.~\ref{sec:mask_repaint}) for precise spatial guidance and Latent Mixture (Sec.~\ref{sec:latent_mixture}) which coherently fuses expert outputs, propagating semantic cues to spatial refinement while mitigating conflicts. Second, our training curriculum is set to be progressive. The model is initially exposed to basic, single-task training data, then graduates to complex multi-task samples. This allows the experts to evolve from generic representations to specialization, mitigating mode collapse and improving generalization over diverse editing behaviors. As such, compared to static-fusion editors like OmniControl-, or DiT-adapter variants~\citep{zhang2023controlnet, zhang2024omnicontrol, peebles2023dit}, CARE-Edit provides selective, condition-aware compute that better resolves the conflicts among text, mask, and reference signals (Sec.~\ref{sec:empirical_analysis}).

We evaluate CARE-Edit on instruction-based (Sec.~\ref{sec:instruction_editing}) and subject-driven (Sec.~\ref{sec:subject_editing}) settings, covering diverse tasks such as object erasure, replacement, text-driven edits, and localized style transfer. CARE-Edit achieves strong results, improving edit faithfulness, boundary cleanliness, and identity/style preservation over unified editors and swap-style pipelines~\citep{brock2019biggan, xu2022styleswin, park2019spade, liu2021fusedreamtrainingfreetexttoimagegeneration}. Training dynamics and task-expert relationsip analysis demonstrate the effectiveness of condition-aware processing to reduce task interference. 

Our contributions can thus be summarized as follows:
\begin{itemize}
    \item We identify static, model capacity-agnostic fusion of conditions as a major source of conflicts in existing image editors, and thus propose CARE-Edit that employs condition-aware routing of experts for dynamic compute allocation.

    \vspace{-0.35em}
    \item We introduce three complementary modules to maximize CARE-Edit's fidelity: (i) Mask Repaint to refine arbitrary, user-defined masks for precise spatial control; (ii) Latent Mixture to coherently aggregate expert outputs; and (iii) Routing Select that implements top-$K$ activation, ensuring only the most relevant experts process each specific token.

    \vspace{-0.35em}
    \item We demonstrate competitive performance of CARE-Edit on diverse editing tasks, with empirical analysis showcasing that dynamic, condition-aware experts are effective for resolving conflicts in multi-condition image editing.
\end{itemize}

\section{Related Work}

\paragraph{Instruction-based Editing.}
Text-driven diffusion image editing can be classified into two groups.
(i) Task-specific methods~\citep{zhang2024magicbrushmanuallyannotateddataset,hui2024hqedithighqualitydatasetinstructionbased,ultraedit}: global refinement via re-sampling (SDEdit~\citep{meng2022sdedit}); semantic steering via attention or PnP~\citep{hertz2022prompt2prompt} and NTI~\citep{mokady2023nulltext}; instruction-following editing~\citep{brooks2023instructpix2pix}; localized editing via mask discovery~\citep{couairon2023diffedit} or conditional control~\citep{zhang2023controlnet,zhang2024omnicontrol}; Recent pipelines like EMU-Edit~\citep{sheynin2024emu}. These methods excel under single-signal guidance but struggle with multi-source, conflicting constraints due to static fusion.
(ii) Unified editors~\citep{meng2024instructgiegeneralizableimageediting,mige2024,xu2025incontextbrushzeroshotcustomized, zhang2025icedit}: systems that consolidate heterogeneous objectives within a single interface, exemplified by ACE++~\citep{ace2024,mao2025aceinstructionbasedimagecreation}, OmniGen2~\citep{omnigen2024,wu2025omnigen2}, and AnyEdit~\citep{anyedit}. Despite the progress, resolving multi-condition conflicts remains a significant challenge in contextual image editing.

\vspace{-1.00em}
\paragraph{Subject-driven Editing.}
Subject-based image editing spans embedding-based or adapter-based methods (\eg, DreamBooth~\citep{dreambooth}, LoRA~\citep{hu2022lora}) that learn subject/style concepts. It risks overfitting or unintended editing outside target regions. Recent methods extend reference conditioning, including BLIP-Diffusion~\citep{blipdiffusion}, OmniControl~\citep{zhang2024omnicontrol}, UNO~\citep{uno}, and unified editors such as OmniGen2~\citep{omnigen2024,wu2025omnigen2}; earlier subject-centric editors (\eg, MimicBrush~\citep{chen2024zeroshotimageeditingreference}, AnyDoor~\citep{chen2024anydoorzeroshotobjectlevelimage}) also explore appearance transfer with varying locality and generalization. In contrast, we treat reference guidance as a conditional competency handled by specialized experts.

\vspace{-1.00em}
\paragraph{Mixture-of-Experts for Image Editing.}
Sparse MoE models scale capacity via routed specialization~\citep{shazeer2017outrageously,lepikhin2020gshard,fedus2022switch,roller2021hash}, and the recent diffusion MoE (\eg, EC-DiT~\citep{sun2025ecditscalingdiffusiontransformers} with adaptive expert-choice routing) shows timestep-aware token routing is effective. Unlike these application of homogenous experts, we employ heterogeneous experts (text, mask, reference, and base) to solve multi-condition conflicts. A timestep-aware router selectively activates these experts along denoising trajectory, which adaptively allocates model capacity for different conditions in contexual image editing.

\begin{figure*}[t]
    \centering
    \includegraphics[width=\textwidth]{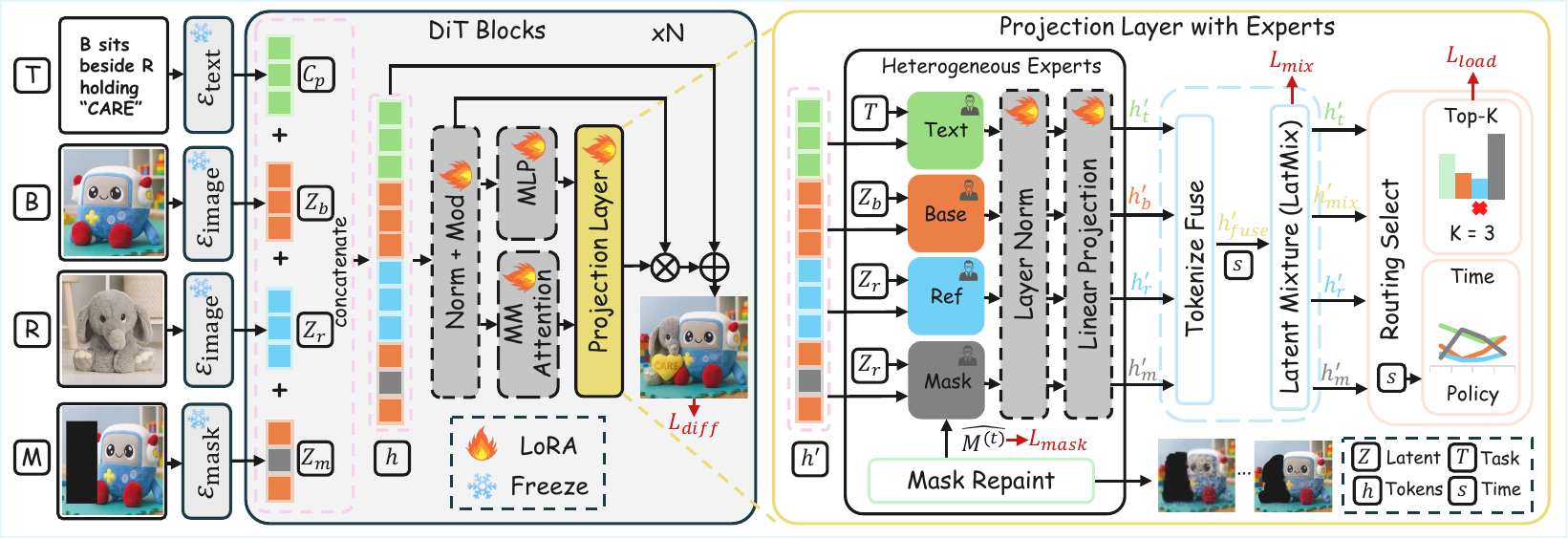}
    \vspace{-1.40em}
    \caption{CARE-Edit introduces condition‑aware specialized experts within the frozen DiT backbone.
Given multimodal conditions, 
inputs are tokenized and projected to heterogeneous expert branches.
The router assigns confidence scores and selects the top‑$K$ experts to process each token.
Expert outputs are normalized, modulated, and fused through the Latent Mixture module,
yielding denoised representations $\mathbf{h}'$ refined by Mask Repaint.
Only lightweight adapters, the router, and fusion layers are trainable.
This enables CARE-Edit to dynamically allocate computation, mitigates conflicts between heterogeneous conditions (\eg, text vs mask) and enables high-fidelity, coherent edits.}
    \label{fig:architecture}
    \vspace{-0.60em}
\end{figure*}

\section{Methodology}

We propose CARE-Edit, a diffusion-based editor that routes computation to condition-aware experts. As aforementioned, the key challenge in unified image editing is task interference, where conflicting conditions from text, masks, and reference images lead to artifacts like style bleeding or identity loss. Unlike static-fusion approaches that process all conditions with a shared backbone, CARE-Edit performs fine-grained condition-aware routing over a set of heterogeneous experts, each specialized in processing a particular modality or function. They communicate through cross-modal interactions, enabling the model to selectively integrate multi-condition information across the denoising process. This \textit{specialize-then-fuse} manner allows CARE-Edit to dynamically allocate computation, prioritize relevant modalities, and thus potentially mitigate conflicts between competing edit instructions.

\subsection{Preliminaries}
\label{sec:preliminaries}

As shown in Figure \ref{fig:overview}, different image editing paradigms can be categorized by the input modalities used for conditioning.  
Let $\mathbf{T}$ denote the text prompt, $\mathbf{B}\!\in\!\mathbb{R}^{H\times W\times3}$ the base image,  $\mathbf{R}\!\in\!\mathbb{R}^{H\times W\times3}$ the reference image, and $\mathbf{M}\!\in\![0,1]^{H\times W}$ a binary or soft mask indicating editable regions. Four representative settings are considered:  
\textit{Instruction-based} editing conditions on $(\mathbf{T}, \mathbf{B})$, where textual guidance directs modifications to the base image;  
\textit{Subject-based} ones condition on $(\mathbf{B}, \mathbf{R})$ to maintain identity or style;  
\textit{Contextual} editing incorporates $(\mathbf{T}, \mathbf{B}, \mathbf{R})$ to confine edits to specific regions. CARE-Edit unifies these modalities $(\mathbf{T}, \mathbf{B}, \mathbf{R}, \mathbf{M})$ within a single framework for flexible, condition-aware generation.

\vspace{-1.10em}
\paragraph{Modality Encoders.}  
Each input modality is first mapped to a latent token sequence by a specialized, frozen encoder.  
The text prompt is processed by a Text Encoder $\mathcal{E}_{\text{text}} (\cdot)$, producing contextual embeddings 
$\mathbf{C}_p = \mathcal{E}_{\text{text}}(\mathbf{T}) \in \mathbb{R}^{N_t \times d}$,  
where $N_t$ denotes the number of text tokens and $d$ indicates the feature dimension of each token embedding.  
The Image Encoder $\mathcal{E}_{\text{image}} (\cdot)$ (\eg, DINO~\citep{caron2021emerging} or VAE~\citep{kingma2013auto}) extracts latent representations 
$\mathbf{Z}_b = \mathcal{E}_{\text{image}}(\mathbf{B})$ for the base image and $\mathbf{Z}_r = \mathcal{E}_{\text{image}}(\mathbf{R})$ for the reference image.  
A Mask Encoder $\mathcal{E}_{\text{mask}} (\cdot)$ converts the spatial mask into aligned latent tokens 
$\mathbf{Z}_m = \mathcal{E}_{\text{mask}}(\mathbf{M})$, which enables explicit region control.

\vspace{-1.10em}
\paragraph{Latent Composition and DiT Backbone.}  
These latents are projected to a shared embedding space and concatenated:
\[
\mathbf{h}_0 = [\mathbf{C}_p; \mathbf{Z}_b; \mathbf{Z}_r; \mathbf{Z}_m] \in \mathbb{R}^{N \times d},
\quad \text{where} \; N = N_t + 3N_v,
\]
where $N_v$ indicates the number of visual tokens from each image- or mask-related latent.  
This unified token sequence is then propagated through the diffusion transformer (DiT) backbone \cite{peebles2023dit} which is parameterized by $\theta$:
\[
\mathbf{h}_{t} = \mathcal{D}_{\mathrm{DiT},\,\theta}\!\left(\mathbf{h}_{0},\, t\right),
\]
where $t$ denotes the diffusion timestep.

\vspace{-1.10em}
\paragraph{Fine‑Tuning and Optimization.}  
We apply LoRA‑style fine‑tuning \cite{hu2022lora} on a pretrained \textsc{Flux} diffusion model~\citep{flux2024} to adapt the DiT backbone~\cite{peebles2023dit} to multi‑modal conditioning while preserving pretrained generative priors.  
The model is optimized via the standard denoising diffusion objective as:
\[
\mathcal{L}_{\text{diff}} =
\mathbb{E}_{t,\mathbf{h}_0,\boldsymbol{\epsilon}}
\big[
\|\boldsymbol{\epsilon} - \boldsymbol{\epsilon}_{\theta}(\mathbf{h}_{t}, t)\|^2
\big],
\]
where $\boldsymbol{\epsilon}\!\sim\!\mathcal{N}(0,\mathbf{I})$ denotes the Gaussian noise.  
This formulation unifies all modalities into a shared latent representation, laying the groundwork for the condition-aware expert routing discussed in the following sections.

\subsection{Overview of CARE-Edit}
\label{sec:method_overview}

As illustrated in Figure~\ref{fig:architecture}, CARE-Edit is built around three core components:
(1)~\textit{Mask Repaint}, which refines the coarse, user‑provided masks into spatially accurate ones guided by reference geometry;
(2)~\textit{Latent Mixture}, which fuses the reference and base features within the masked region with per-token and per-timestep awareness; and
(3)~\textit{Routing Select}, which dynamically activates the top‑$K$ modality experts to allocate task‑specific capacity during the diffusion process.
Together, these enable semantically consistent and spatially coherent multimodal editing in a unified DiT framework.

\vspace{-1.10em}
\paragraph{LoRA Fine‑Tuning.}  
Following the previous studies like OmniControl~\citep{zhang2024omnicontrol}, we integrate LoRA adapters~\cite{hu2022lora} into the self-attention and projection layers of each DiT block while keeping the pretrained encoders frozen.
This configuration, as shown in Figure~\ref{fig:architecture}, enables flexible multi-modal modulation and expert routing without fullly retraining the model.
Within each DiT block, we denote by $\mathbf{h}^{\,'}\!\in\!\mathbb{R}^{N\times d}$ the latent token set immediately before the Projection Layer.

\vspace{-1.10em}
\paragraph{Specialized Experts.} CARE‑Edit employs four heterogeneous experts corresponding to the text, mask, reference, and base modalities.
The text expert performs semantic reasoning and object synthesis through cross‑attention with text tokens.
The mask expert focuses on spatial precision and boundary refinement guided by the edit mask.
The reference expert learns identity‑ and style‑consistent transformations from reference features.
The base expert enforces global coherence and background consistency.
Each expert functions as a lightweight adapter embedded in a DiT block, introducing modality‑specific inductive bias with minimal additional parameters. $\mathbf{h}^{\,'}_b$, $\mathbf{h}^{\,'}_r$, $\mathbf{h}^{\,'}_t$, and $\mathbf{h}^{\,'}_m$  
represent the tokens from the base, reference, text, and mask experts, respectively.

\vspace{-1.10em}
\paragraph{Unified Dimensionality.} 
Despite differences in structure, all experts share the backbone's input and output dimensionality $d$, with each defined as $f^{e}\!:\mathbb{R}^{N\times d}\!\rightarrow\!\mathbb{R}^{N\times d}$.
The expert outputs pass through a \texttt{LayerNorm} followed by a \texttt{Linear} projection layer to maintain feature‑scale consistency, which is proven significant for residual stability during training.

\vspace{-1.10em}
\paragraph{Token‑wise Top‑K Routing.} After obtaining the latent tokens $\mathbf{h}'$ from each DiT block, CARE‑Edit performs token‑wise routing to determine which modality experts should process each token. 
For every token $\mathbf{h}'_i$, a router computes a probability distribution $\pi_{i,e}$ over the four experts by combining local content features and global task context. 
Following the Mixture‑of‑Experts formulation in \cite{shazeer2017outrageously,fedus2022switch}, 
a token‑specific key $\mathbf{k}_i = W_k \mathbf{h}'_i$ encodes local information, while a global conditioning query 
$\mathbf{q} = \phi(\mathbf{T})$ summarizes the current editing objective, where $\mathbf{T}$ denotes the task‑condition embedding 
representing removal, replacement, text‑driven edits, or localized style transfer. 
The routing temperature $\tau$ is gradually annealed during training, and router logits are smoothed with an exponential moving average to reduce variance, which stabilizes dynamic expert selection. 
As such, the expert-specific logits then can be computed as:
\begin{equation}
\alpha_{i,e} = \text{MLP}_e([\mathbf{k}_i \Vert \mathbf{q}]) + b_e,
\end{equation}
where $\alpha_{i,e}$ denotes the pre‑softmax activation (logit) for expert $e$, and $b_e$ is a learnable bias that captures prior expert preferences. 
The softmax‑based routing probabilities and Top‑$K$ sparsification are jointly expressed as:
\begin{equation}
\tilde{\pi}_{i,e} =
\frac{
    \exp(\alpha_{i,e}/\tau) \,
    \mathbf{1}[e \!\in\! \mathcal{S}_i]
}{
    \sum_{j \in \mathcal{S}_i} \exp(\alpha_{i,j}/\tau)
},
\label{eq:routing}
\end{equation}
where $\mathcal{S}_i$ is the index set of the $K$ experts with the highest scores for token $i$. 
In practice, $K$ is set to 3, achieving a favorable balance between representational diversity and computational efficiency. 
This scheme allows each token to adaptively attend to the most relevant experts (with specific modalities) underthe  spatial–semantic–task joint guidance.

\vspace{-1.10em}
\paragraph{Residual Aggregation and Stability.} At this stage, each expert refines $\mathbf{h}'$ through their specific compact adapter as:
\begin{align}
\mathbf{h}^{\,'}_t &= \mathbf{h}' + A^ {\text{text}}\!\left(\text{CrossAttn}(\mathbf{h}', \mathbf{T})\right),\\
\mathbf{h}^{\,'}_b &= \mathbf{h}' + A^{\text{base}}\!\left(\text{CrossAttn}(\mathbf{h}', Z_b)\right),\\
\mathbf{h}^{\,'}_r &= \mathbf{h}' + A^{\text{ref}}\!\left(\text{FiLM}(\mathbf{h}',Z_r)\right),\\
\mathbf{h}^{\,'}_m &= \mathbf{h}' + A^{\text{mask}}\!\left(\text{Conv}\!\left(\mathbf{h}' \odot \text{Up}(Z_m\odot \hat{M}^{(t)})\right)\right),
\end{align}
where $A^{(\cdot)}$ indicates the lightweight convolutional adapters. 
Cross‑attention follows the conditioning paradigm of latent diffusion~\citep{rombach2022latent,peebles2023dit, wang2025moditlearninghighlyconsistent}. 
To prevent routing collapse, a fixed fraction $\lambda_{\text{shared}}$ of tokens is always routed through a shared expert, which preserves the representation continuity:
\begin{equation}
\tilde{\pi}^{+}_{i,e} =
(1-\lambda_{\text{shared}})\tilde{\pi}_{i,e}
+ \lambda_{\text{shared}}\mathbf{1}[e=\text{share}],
\label{eq:shared_routing}
\end{equation}
and the final residual aggregation is computed as:
\begin{equation}
\mathbf{h}''_i
= \mathbf{h}'_i
+ \sum_e \tilde{\pi}^{+}_{i,e}\big(f^{e}(\mathbf{h}'_{i,e})-\mathbf{h}'_i\big).
\end{equation}
This convex residual fusion aims to stabilize the gradient propagation and thus ensure balanced expert influence.

\vspace{-1.00em}
\paragraph{Overall Training.} CARE‑Edit is trained jointly with the diffusion reconstruction objective. 
To encourage balanced expert utilization and stabilize the routing, 
we add a load‑balancing regularizer following prior MoE work~\citep{sun2025ecditscalingdiffusiontransformers}:
\begin{equation}
\mathcal{L}_{\text{load}}
=\sum_e
\left(
\frac{1}{N}\sum_i \pi_{i,e}
- \frac{1}{|E|}
\right)^{2}.
\end{equation}

\subsection{Mask Repaint}
\label{sec:mask_repaint}

The user-defined masks $\mathbf{M}$ could misalign with object boundaries, thereby causing artifacts and color bleeding.
To address this, \emph{Mask Repaint} module refines $\mathbf{M}$ at each diffusion step $t$ by exploiting geometric correspondence between current latent and reference features. 
It predicts a soft, boundary-aware mask that adapts to object contours and thus promotes smooth transitions between edited and preserved regions.

\vspace{-1.10em}
\paragraph{Reference–guided Refinement.}
At timestep~$t$, the module takes the current latent~$\mathbf{h}'^{t}$, the reference encoding~$Z_r$, and the previous‑step predicted mask latent~$\hat{M}^{(t-1)}$.  
A residual mask field~$\Delta\mathbf{m}$ is estimated from concatenated features:
\begin{equation}
\Delta\mathbf{m} =
\sigma\!\left(
W_2\,\text{Conv}\!\big(
[\mathbf{h}'^{(t)} \,\Vert\,\text{Up}( Z_r )\,\Vert\, \text{Up}(\hat{M}^{(t-1)})]
\big)
\right),
\label{eq:mask_residual_refine}
\end{equation}
where $W_m$ is a projection layer, and $\sigma(\cdot)$ denotes sigmoid activation.  
The refined soft mask is then given by:
\begin{equation}
\hat{M}^{(t)} =
\text{clip}\!\big(
\hat{M}^{(t-1)} + \Delta\mathbf{m},\, 0,\, 1
\big),
\label{eq:mask_refined}
\end{equation}
which adaptively aligns to boundaries in both $\mathbf{h}'^{t}$ and $Z_r$.  
Since the update operates in latent space, it yields spatially coherent masks without explicit pixel‑wise supervision.

\vspace{-1.10em}
\paragraph{Integration with Diffusion.}
The refined mask~$\hat{M}^{(t)}$ is fed back into the routing process of the next CARE-Edit diffusion block, modulating the \textit{mask} and \textit{base} experts.  
In practice, it serves as the dynamic spatial prior within Eq. (6), updating the mask interaction
$\text{Up}(\hat{M}^{(t)})$ that gates token features at the subsequent denoising step.  
This iterative refinement enables progressively sharper boundary control and seamless region blending across diffusion steps.

\vspace{-1.10em}
\paragraph{Training.}
Mask Repaint is trained jointly with the diffusion objective using a boundary‑consistency loss \cite{kervadec2019boundary}:
\begin{equation}
\mathcal{L}_{\text{mask}} =
 \bigl\|\nabla\hat{M}^{(t)} - \nabla M_{\text{gt}}\bigr\|_1
 + \lambda_{\text{smooth}}
   \bigl\|\nabla^2\hat{M}^{(t)}\bigr\|_1,
\label{eq:mask_loss_refine}
\end{equation}
where $M_{\text{gt}}$ is a ground‑truth or pseudo mask.  
The first term enforces accurate boundary localization, while the smoothness term suppresses spurious oscillations, producing clean and temporally stable mask evolution throughout denoising.

\definecolor{BestColor}{RGB}{26,140,26}    
\definecolor{SecondColor}{RGB}{0,90,200}   

\begin{figure*}[t]
    \centering
    \includegraphics[width=\textwidth]{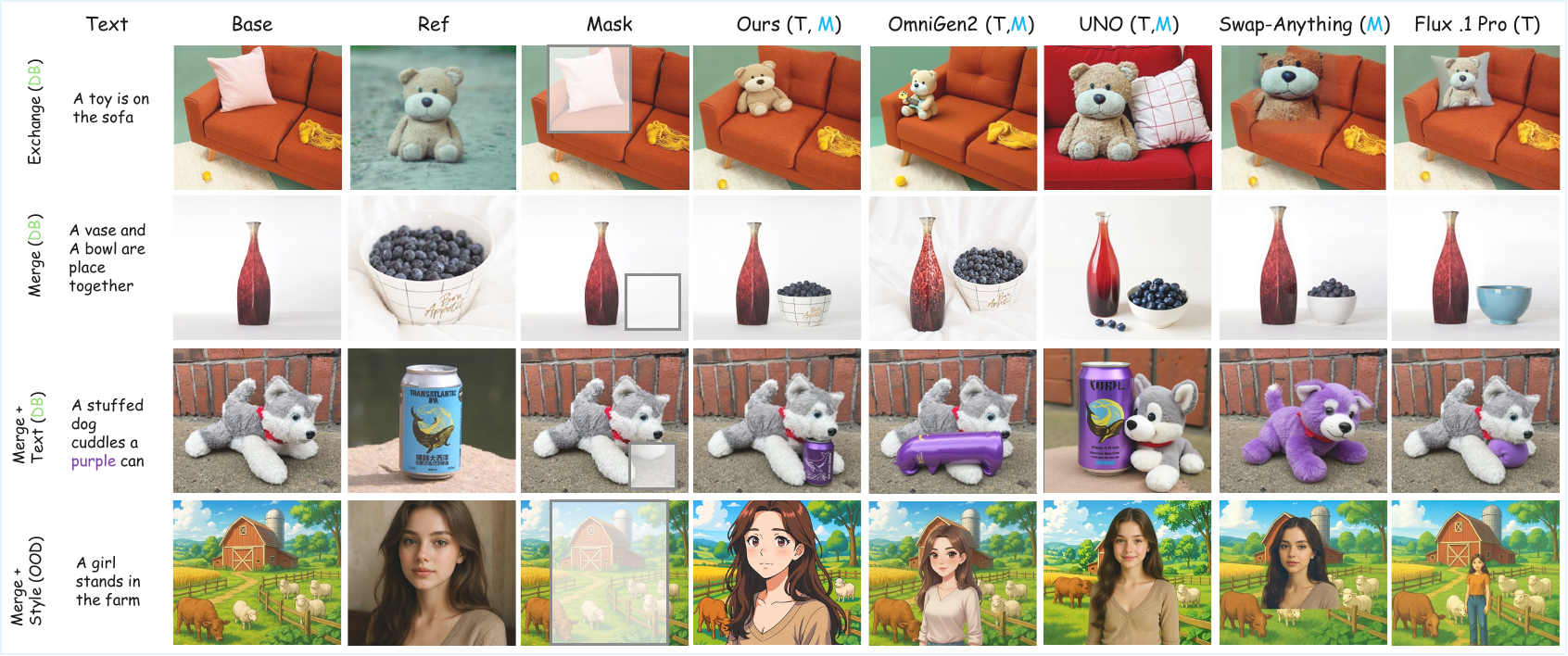}
    \vspace{-1.90em}
    \caption{Qualitative comparison of subject-driven contextual editing. {\color{BestColor}\textbf{Green}} annotations indicate data belonging to the benchmark, while {\color{SecondColor}\textbf{Blue}} annotations denote whether mask inputs are provided. CARE-Edit ensures the preservation of subject identity and coherent contexts.}
    \label{fig:maincompare}
\end{figure*}

\subsection{Latent Mixture}
\label{sec:latent_mixture}

The output of above specialized experts must be coherently aggregated to the final denoised latent.  
Naive concatenation or averaging could blur local details and weaken modality‑specific information. As such, we employ a \emph{Latent Mixture} module that performs per-token and per-timestep processing based on routing confidence and contextual cues.

\vspace{-1.10em}
\paragraph{Token-wise Fusion.}
At each diffusion step, let $\mathbf{h}^{\,'}_e$ denote the output of active experts $e$. From the normalized routing weights $\tilde{\pi}^{+}_{i,e}$ (Eq.~\ref{eq:shared_routing}), we can compute a probability distribution map $\mathbf{w}_e \!\in\! \mathbb{R}^{N\times d}$ for each expert satisfying $\sum_e \mathbf{w}_e = \mathbf{1}$. The fused latent $\mathbf{h}^{\,'}_{fuse}$ is then obtained through a convex combination of expert outputs, as:
\begin{equation}
\mathbf{h}^{\,'}_{fuse} = \sum_{e} \mathbf{w}_e \odot \mathbf{h}^{\,'}_{e},
\end{equation}
where each channel of $\mathbf{h}^{\,'}_{fuse}$ integrates text, semantic, and mask cues according to the router’s attention pattern.

\vspace{-1.10em}
\paragraph{Timestep-adaptive Mixture.}
To maintain global coherence, we blend $\mathbf{h}'_{fuse}$ with the base expert's output via a learned, timestep-dependent gate. A modulation network computes an adaptive gating coefficient $\gamma(\mathbf{h_b}, s)$ from the base‑expert feature and the timestep embedding as:
\begin{equation}
\gamma = \sigma\left(W{\gamma}[\text{GAP}(\mathbf{h_{b}}) || \psi(s)]\right),
\end{equation}
where $\text{GAP}$ denotes global average pooling and $\psi(s)$ is a sinusoidal timestep encoding.
The final latent integrates global background and context‑dependent foreground edits:
\begin{equation}
\mathbf{h}^{\,'}_{mix} = (1-\gamma)\mathbf{h}^{\,'}_{fuse} + \gamma\mathbf{h}^{\,'}_{b}
\end{equation}
This anchors output to the base's global structure while sharpening attention in regions driven by semantic/mask signals.

\vspace{-1.10em}
\paragraph{Training Objective.}
The Latent Mixture module is optimized jointly with the overall diffusion objective.  
To encourage spatial coherence of the mixture maps, 
we introduce a total‑variation regularizer from previous studies~\citep{johnson2016perceptual} as:
\begin{equation}
\mathcal{L}_{\text{mix}}
=
\lambda_{\text{tv}}
\sum_{e}\|\nabla\mathbf{w}_e\|_1,
\label{eq:tv_loss}
\end{equation}
where $\nabla\mathbf{w}_e$ computes finite‑difference spatial gradients of the weight map, and $\lambda_{\text{tv}}$ controls the trade‑off between mixture smoothness and routing flexibility. Larger $\lambda_{\text{tv}}$ enforces stronger spatial continuity in selection,
while a smaller one allows sharper expert transitions for fine‑grained boundaries.

\subsection{Training Loss}

CARE-Edit is trained end‑to‑end by combining the diffusion reconstruction loss (Sec.~\ref{sec:preliminaries}) with the three aforementioned auxiliary regularizers:
a load‑balancing loss $\mathcal{L}_{\text{load}}$
(Sec.~\ref{sec:method_overview}), a mask‑boundary consistency loss $\mathcal{L}_{\text{mask}}$
(Sec.~\ref{sec:mask_repaint}), and a latent‑mixture smoothness loss $\mathcal{L}_{\text{mix}}$ (Sec.~\ref{sec:latent_mixture}) as:
\begin{equation}
\mathcal{L}_{\text{CARE}}
=
\mathcal{L}_{\text{diff}}
+ \lambda_{\text{load}}\,\mathcal{L}_{\text{load}}
+ \lambda_{\text{mask}}\,\mathcal{L}_{\text{mask}}
+ \lambda_{\text{mix}}\,\mathcal{L}_{\text{mix}}
\label{eq:lcare}
\end{equation}

Throughout training, only the expert adapters, router parameters, and lightweight fusion modules are optimized,
while the pretrained denoising backbone remains frozen.

\subsection{Training Dataset}
Our training data is drawn from MagicBrush~\citep{zhang2024magicbrushmanuallyannotateddataset} and OmniEdit~\cite{wei2025omnieditbuildingimageediting}, supplemented with selections from UNO~\cite{uno} to broaden object categories and AnyEdit~\citep{anyedit} for edit types. To target complex contextual editing with identity preservation, we curate a $20K$ multi-paired corpus (Image+Mask+Prompt, optional reference) from Subjects200K~\citep{zhang2024omnicontrol}. It offers samples focusing on diverse objects and humans within shared backgrounds. For mask supervision, we introduce a pipeline to generate precise multi-paired masks for high-quality spatial control. Please refer to Appendix~\ref{app:dataset} for more details.

\section{Experiments}
\label{sec:exp}

\begin{table*}[h]
\centering
\begin{minipage}{\textwidth}
\renewcommand{\arraystretch}{0.95}
\caption{Quantitative results on EMU-Edit~\citep{sheynin2024emu} and MagicBrush~\cite{zhang2024magicbrushmanuallyannotateddataset} test sets. All included editors are classified into \textit{Task-specific} and \textit{Unified} models.
Best and second-best results per metric are marked in \textbf{bold} and \underline{underline}.
$\uparrow (\downarrow)$ indicates higher (lower) values are better.}
\label{tab:emu_magic_comparison}
\vspace{-0.70em}
\resizebox{1.0\textwidth}{!}{
%
%
\begin{tabular}{llll *{8}{c}}
\toprule[1.2pt]
\multirow{2}{*}{Category} & \multirow{2}{*}{Method} & \multirow{2}{*}{Venue} & \multirow{2}{*}{Backbone} 
& \multicolumn{4}{c}{EMU-Edit~\citep{sheynin2024emu} Test} & \multicolumn{4}{c}{MagicBrush~\cite{zhang2024magicbrushmanuallyannotateddataset} Test} \\
\cmidrule(lr){5-8} \cmidrule(lr){9-12}
 & & & & {CLIPim $\uparrow$} & {CLIPout $\uparrow$} & {L1 $\downarrow$} & {DINO $\uparrow$}
   & {CLIPim $\uparrow$} & {CLIPout $\uparrow$} & {L1 $\downarrow$} & {DINO $\uparrow$} \\
\midrule

\multirow{4}{*}{\textbf{Task-specific Models}}
 & PnP~\citep{hertz2022prompt2prompt} & CVPR'23 & SD1.5 &
   0.521 & 0.089 & 0.089 & 0.304 &
   0.568 & 0.101 & 0.289 & 0.220 \\
 & Null-Text~\citep{mokady2023nulltext} & CVPR'23 & SD1.5 &
   0.761 & 0.236 & 0.075 & 0.678 &
   0.752 & 0.263 & 0.077 & 0.664 \\
 & InstructPix2Pix~\citep{brooks2023instructpix2pix} & CVPR'23 & SD1.5 &
   0.834 & 0.219 & 0.121 & 0.762 &
   0.837 & 0.245 & 0.093 & 0.767 \\
 & EMU-Edit~\citep{sheynin2024emu} & CVPR'24 & -- &
   0.859 & 0.231 & 0.094 & 0.819 &
   \underline{0.897} & 0.261 & \underline{0.052} & 0.879 \\
\midrule

\multirow{6}{*}{\textbf{Unified Models}}
 & FLUX.1 Fill~\citep{flux2024} & HuggingFace & FLUX.1 Fill &
   0.663 & 0.205 & 0.176 & 0.674 &
   0.725 & 0.235 & 0.208 & 0.661 \\
 & ACE (ACE++)~\citep{ace2024,mao2025aceinstructionbasedimagecreation} & ICLR'26 & FLUX.1 Fill &
   0.831 & 0.256 & \textbf{0.073} & 0.802 &
   0.818 & 0.268 & \textbf{0.042} & 0.823 \\
 & OmniGen2~\citep{omnigen2024, wu2025omnigen2} & CVPR'25 & FLUX.1 Dev &
   \underline{0.865} & \underline{0.306} & 0.088 & \underline{0.832} &
   \textbf{0.905} & \underline{0.306} & 0.055 & \underline{0.889} \\
 & AnyEdit~\citep{anyedit} & CVPR'25 & SD1.5 &
   0.866 & 0.284 & 0.095 & 0.812 &
   0.892 & 0.273 & 0.057 & 0.877 \\
 & \textbf{CARE-Edit (Ours)} & -- & FLUX.1 Dev &
   \textbf{0.868} & \textbf{0.313} & \underline{0.082} & \textbf{0.835} &
   0.894 & \textbf{0.324} & \underline{0.052} & \textbf{0.885} \\
\bottomrule[1.2pt]
\end{tabular}}
\end{minipage}
\end{table*}

We test CARE-Edit on three mainstream benchmarks, covering (i) instruction-based editing on EMU-Edit~\citep{sheynin2024emu} and MagicBrush~\citep{zhang2024magicbrushmanuallyannotateddataset}, and (ii) subject-driven contextual editing on recent DreamBench++~\citep{peng2025dreambenchhumanalignedbenchmarkpersonalized}, including its challenging multi-object settings to ensure rigorous evaluation.
We first introduce experimental details and evaluation metrics (Sec.~\ref{sec:exp_setup}), followed by direct comparisons on both instruction-based (Sec.~\ref{sec:instruction_editing}) and contexual (Sec.~\ref{sec:subject_editing}) tasks. We then present ablation studies (Sec.~\ref{sec:ablation}) and empirical analysis (Sec.~\ref{sec:empirical_analysis}) that validate CARE-Edit beyond benchmarking results.

\subsection{Experimental Setup}
\label{sec:exp_setup}

\paragraph{Implementation Details.}
We adopt a DiT-style backbone, interleaving routed layers every 2--3 blocks to balance the overhead and representation power. Each routed layer uses top-$K$ expert routing with $K \in \{2,3,4\}$, with a shared expert activated for stability and global coherence~\citep{lepikhin2020gshard, roller2021hash, fedus2022switch}. The model is first finetuned following Ominicontrol~\cite{zhang2024omnicontrol}, and then jointly optimize with sparse routing to specialize experts. We use AdamW optimizer with a learning rate of $1\times10^{-4}$, a batch size of $16$, and weight decay of $0.01$. All models are trained for $100K$ iterations on $8 \times$ NVIDIA L20 GPUs using a cosine-decay learning rate scheduler.

\vspace{-1.20em}
\paragraph{Curriculum Training.}
We adopt a cross-task curriculum that gradually increases task difficulty, which promotes stable training and guides specialized expert learning. The model is first trained for $40K$ iterations on basic, single-task samples, and then switches to complex multi-task data for the remaining $60K$ iterations. This progressive schedule allows the routing layers to evolve from generic representations to specialized functions, mitigating mode collapse and improving generalization over diverse editing behaviors.

\vspace{-1.20em}
\paragraph{Evaluation Metrics.}
We follow standard protocols to evaluate performance and report image-level subject consistency using DINO-I~\citep{oquab2023dinov2} and CLIP-I~\citep{radford2021learning}, and text–image alignment using CLIP-T~\citep{radford2021learning}. For all three metrics, higher scores indicate better subject fidelity and semantic coherence.

\begin{table}[t]
\centering
\renewcommand{\arraystretch}{0.96}
\caption{
Quantitative results on DreamBench++~\citep{peng2025dreambenchhumanalignedbenchmarkpersonalized}.
The best and second‑best results are marked in \textbf{bold} and \underline{underlined}, respectively.
}
\label{tab:dreambenchpp_half}
\vspace{-0.70em}
\resizebox{0.48\textwidth}{!}{
\begin{tabular}{l *{6}{S[table-format=1.3]}}
\toprule[1.2pt]
\multirow{2}{*}{{Method}} &
\multicolumn{3}{c}{{Single‑Object}} &
\multicolumn{3}{c}{{Multiple‑Object}} \\
\cmidrule(lr){2-4}\cmidrule(lr){5-7}
 & {DINO‑I $\uparrow$} & {CLIP‑I $\uparrow$} & {CLIP‑T $\uparrow$}
 & {DINO‑I $\uparrow$} & {CLIP‑I $\uparrow$} & {CLIP‑T $\uparrow$} \\
\midrule
DreamBooth~\citep{dreambooth}        & 0.552 & 0.544 & 0.301 & 0.359 & 0.495 & 0.305 \\
BLIP‑Diffusion~\citep{blipdiffusion} & 0.610 & 0.649 & 0.293 & 0.462 & 0.592 & 0.289 \\
OmniControl~\citep{zhang2024omnicontrol}      & 0.770 & 0.704 & 0.312 & 0.501 & 0.641 & 0.316 \\
UNO~\citep{uno}                      & 0.782 & 0.713 & 0.304 & 0.508 & 0.649 & 0.303 \\
OmniGen2~\citep{omnigen2024, wu2025omnigen2}             & \underline{0.861} & \underline{0.784} & \underline{0.318}
                                     & \underline{0.560} & \underline{0.713} & \underline{0.319} \\
\midrule
\textbf{CARE‑Edit (Ours)}            & \textbf{0.874} & \textbf{0.792} & \textbf{0.325}
                                     & \textbf{0.568} & \textbf{0.720} & \textbf{0.327} \\
\bottomrule[1.2pt]
\end{tabular}}
\vspace{-0.30em}
\end{table}

\begin{figure}[t]
\centering
\includegraphics[width=\linewidth]{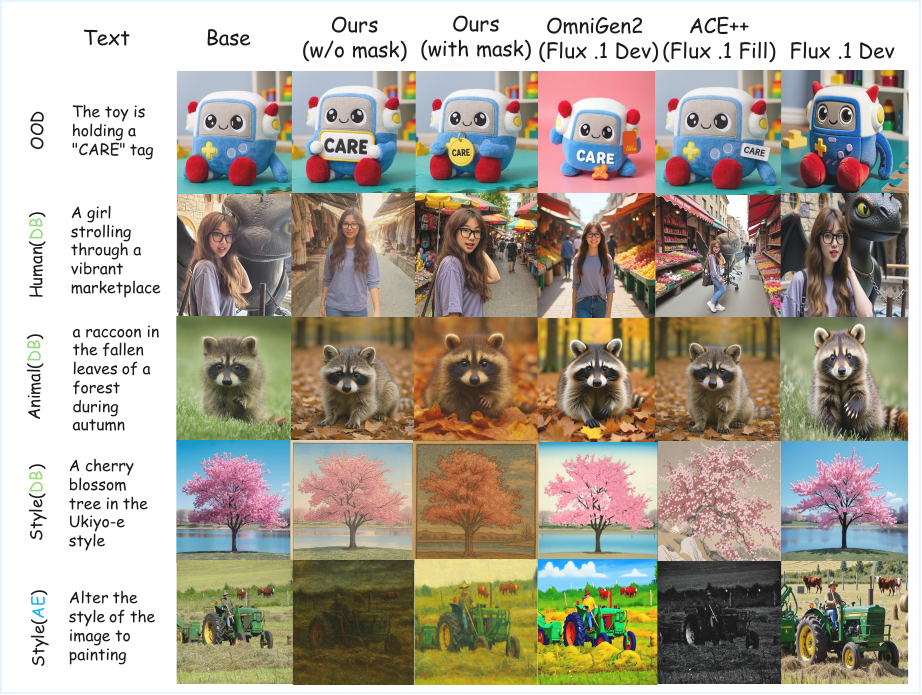}
\vspace{-1.90em}
\caption{Qualitative comparison of instruction-based editing.}
\label{fig:qualitative_analysis}
\vspace{-0.60em}
\end{figure}

\subsection{Instruction-based Image Editing}
\label{sec:instruction_editing}

\vspace{-0.10em}
\paragraph{Baselines.}
We compare CARE-Edit against two classes of baselines: (i) Task-specific methods, including PnP~\citep{hertz2022prompt2prompt}, Null-Text~\citep{mokady2023nulltext}, InstructPix2Pix~\citep{brooks2023instructpix2pix}, and EMU-Edit~\citep{sheynin2024emu}; (ii) Unified editors, including FLUX.1Fill~\citep{flux2024}, ACE/ACE++~\citep{ace2024,mao2025aceinstructionbasedimagecreation}, OmniGen2~\citep{omnigen2024, wu2025omnigen2}, and AnyEdit~\citep{anyedit}. All methods are tested using their official checkpoints or configurations.

\vspace{-1.20em}
\paragraph{Quantitative Results.}
As illustrated in Table~\ref{tab:emu_magic_comparison}, CARE-Edit achieves competitive performance over unified editors on EMU-Edit~\citep{sheynin2024emu} and MagicBrush~\citep{zhang2024magicbrushmanuallyannotateddataset} with approximately $120K$ training data, which is much lower than that of OmniGen2~\citep{wu2025omnigen2}. On EMU-Edit~\citep{sheynin2024emu}, CARE-Edit yields the best CLIPim, CLIPout, and DINO scores, matching or even surpassing task-specific methods. On MagicBrush~\citep{zhang2024magicbrushmanuallyannotateddataset}, CARE-Edit hits the highest CLIPout and DINO scores while keeping a competitive L1. The per-category results demonstrate CARE-Edit's robustness across both local edits (\eg, object removal) and global appearance changes (\eg, style transfer).

\vspace{-1.20em}
\paragraph{Qualitative Results.}
Figure~\ref{fig:qualitative_analysis} visually confirms our quantitative results. CARE-Edit produces cleaner, more instruction-faithful edits with sharper boundaries and fewer artifacts than competing editors. Please view Appendix~\ref{app:qualitative} for details.

\subsection{Subject-driven Contextual Image Editing}
\label{sec:subject_editing}

\paragraph{Baselines.}
We evaluate subject-driven contextual editing against strong personalization and unified editors, including DreamBooth~\citep{dreambooth}, BLIP-Diffusion~\citep{blipdiffusion}, OmniControl~\citep{zhang2024omnicontrol}, UNO~\citep{uno}, and OmniGen2~\citep{omnigen2024, wu2025omnigen2}. All methods use official implementations. We follow the evaluation protocol in UNO~\citep{uno} for the DreamBench++~\citep{peng2025dreambenchhumanalignedbenchmarkpersonalized} multi-object setting.

\vspace{-1.20em}
\paragraph{DreamBench++ Results.}
Table~\ref{tab:dreambenchpp_half} shows that CARE-Edit achieves the best performance across all metrics on DreamBench++ ~\citep{peng2025dreambenchhumanalignedbenchmarkpersonalized} in both single- and multiple-object settings. Notably, it slightly but consistently outperforms the recent strong OmniGen2~\citep{wu2025omnigen2}. This demonstrates that CARE-Edit effectively preserve subject identity and structure, even when accommodating complex multi-object contextual changes.

\vspace{-1.20em}
\paragraph{Qualitative Results.}
Figure~\ref{fig:maincompare} shows that CARE-Edit produces edits with more faithful subject appearance and more coherent backgrounds, reducing artifacts and improving the integration between foreground objects and their surrounding context. Please view Appendix~\ref{app:qualitative} for more details.

\begin{figure}[t]
\centering
\includegraphics[width=\linewidth]{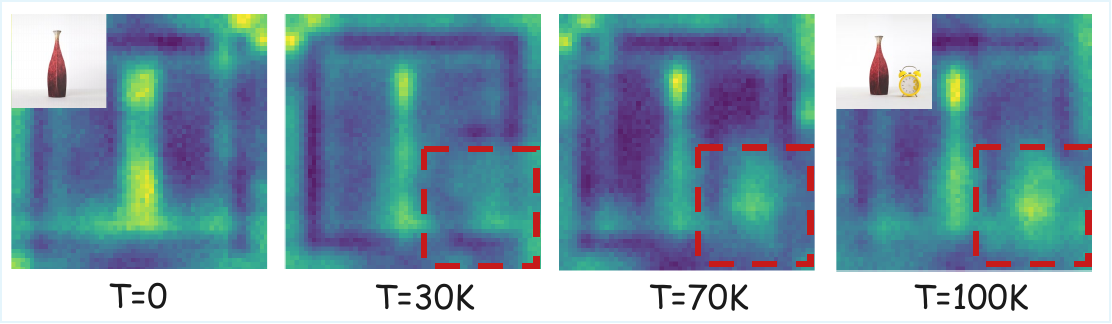}
\vspace{-1.80em}
\caption{Visualization of base expert attention map over iterations.}
\label{fig:expert_attention}
\end{figure}

\begin{figure}[t]
\centering
\includegraphics[width=\linewidth]{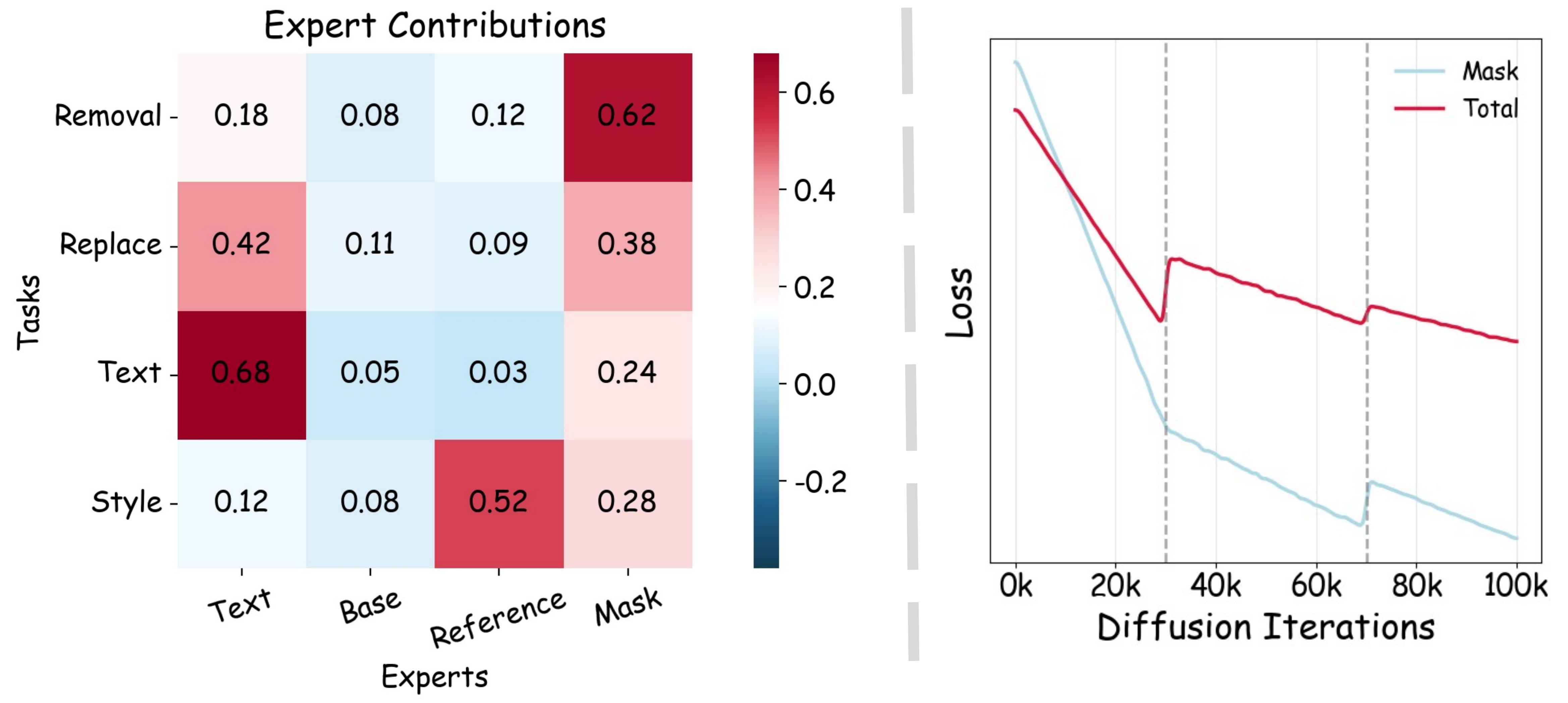}
\vspace{-1.80em}
\caption{Task-Expert Activation Analysis and Loss Curves.}
\label{fig:expert_roles}
\vspace{-0.40em}
\end{figure}

\subsection{Ablation Study}
\label{sec:ablation}

We conduct ablation studies on challenging multiple-object setting, where preserving subject identity and contextual consistency is most difficult. Table~\ref{tab:ablation} isolates the contributions of main components (\eg, expert routing, Latent Mixture, Mask Repaint) and the number of activated experts $K$. 

\vspace{-1.20em}
\paragraph{Impact of Core Components.}
Removing expert routing (w/o Experts) causes large performance drop, indicating that dynamically routing tokens to specialized experts is crucial for handling diverse editing behaviors. Disabling Latent Mixture (w/o Latent Mixture) and Mask Repaint (w/o Mask Repaint) also incur degrades, underscoring their vital roles in aggregating expert outputs and achieving precise edits.

\vspace{-1.20em}
\paragraph{Impact of $K$ in Routing.}
We observe that setting $K=3$ yields optimal results. Using fewer experts may under-express the model while more experts slightly hurts performance, which we attribute to reduced expert specialization. Note that the variations between $K=2, 3,4$ are relatively minimal, which demonstrates the robustness of CARE-Edit.

\begin{table}[t]
\centering
\small   
\renewcommand{\arraystretch}{0.96}
\setlength{\tabcolsep}{4.5pt} 
\caption{
\textbf{Ablation studies of CARE-Edit.}
$\uparrow$ indicates higher values are better.
The best results per metric are highlighted in \textbf{bold}.
}
\label{tab:ablation}
\vspace{-0.70em}
\resizebox{0.43\textwidth}{!}{
\begin{tabular}{lccc}
\toprule[1.2pt]
{Variant} & {DINO‑I $\uparrow$} & {CLIP‑I $\uparrow$} & {CLIP‑T $\uparrow$} \\
\midrule
w/o Experts             & 0.485 & 0.652 & 0.296 \\
w/o Latent\,Mixture     & 0.509 & 0.678 & 0.301 \\
w/o Mask\,Repaint       & 0.523 & 0.693 & 0.304 \\
$K=2$                   & 0.541 & 0.707 & 0.312 \\
$K=4$                   & 0.562 & 0.716 & 0.325 \\
Full Model (K=3)                  & \textbf{0.568} & \textbf{0.720} & \textbf{0.327} \\
\bottomrule[1.2pt]
\end{tabular}
}
\vspace{-0.30em}
\end{table}

\subsection{Empirical Analysis}
\label{sec:empirical_analysis}

\paragraph{Task-Expert Activation.}
To investigate the relative contributions of experts for different tasks, we analyze their activation patterns across four distinct tasks. As shown in Figure~\ref{fig:expert_roles}, different tasks exhibit clear, distinct demands for specific conditions/modalities, which are met by corresponding expert specialization. We observe two key patterns: (i) \textit{Base} Expert maintains robust activation across all tasks, ensuring consistent global representation of the source image. (ii) Other experts exhibit clear specialization: \textit{Mask} Expert dominates structure‑aware edits (\eg, removal and replacement), while the \textit{Reference} Expert is heavily activated during style transfer to maintain fidelity. This dynamic, task-aware activation highlights the limitations of static fusion, which struggle to adaptively allocate resources for multi-condition information. It also demonstrates that CARE-Edit achieves the dynamic multi-condition processing it was designed for.

\vspace{-1.20em}
\paragraph{Expert Latent Attention.}
We visualize \textit{Base} Expert’s latent attention maps to understand how it processes the source image over time. Figure~\ref{fig:expert_attention} shows a clear evolution: In mid-phase ($30K$–$100K$ steps), the updated mask information is injected via Latent Mixture and inter-expert interactions. This leads to progressively sharper, structured attention on masked regions. Interestingly, in the late stage, \textit{Base} Expert evolves beyond merely copying spatial signals and begins to semantically refine masked areas, adjusting its focus based on the editing intent rather than just spatial cues. The training dynamics (Figure~\ref{fig:expert_roles}, right) corroborate this finding: the mask loss decreases in tandem with the total loss, indicating the model is successfully learning to integrate mask-aware information into a semantically-meaningful representation.

\section{Conclusion and Discussion}
We present CARE-Edit that addresses multi-condition conflicts by employing heterogeneous experts with efficient routing for versatile, high-fidelity image editing. It improves controllability via masks and references and scales with modest overhead. We will release our code, models, and the dataset to facilitate research in controllable, multimodal editing.

\vspace{-1.10em}
\paragraph{Limitations and Future Work.}
There are several limitations in this work: (1) The additional hyperparameters (\eg, top-$K$) inherent to CARE-Edit. (2) While the expert set covers most common tasks and modalities, we aim to explore extending it to handle broader edit types in our future work, potentially through dynamic expert loading or expansion.

{
    \small
    \bibliographystyle{ieeenat_fullname}
    \bibliography{main}

@String(CVPR= {IEEE Conf. Comput. Vis. Pattern Recog.})

@String(ICCV= {Int. Conf. Comput. Vis.})

@String(ECCV= {Eur. Conf. Comput. Vis.})

@String(NIPS= {Adv. Neural Inform. Process. Syst.})

@String(ACMMM= {ACM Int. Conf. Multimedia})

@String(ICLR = {Int. Conf. Learn. Represent.})

@String(AAAI = {AAAI})

@String(CVPR  = {CVPR})

@String(ICCV  = {ICCV})

@String(ECCV  = {ECCV})

@String(NIPS  = {NeurIPS})

@String(ACMMM = {ACM MM})

@String(ICLR  = {ICLR})

@inproceedings{ho2020ddpm,
  title={Denoising Diffusion Probabilistic Models},
  author={Ho, Jonathan and Jain, Ajay and Abbeel, Pieter},
  booktitle={NIPS},
  year={2020}
}

@inproceedings{song2020score,
  title={Score-Based Generative Modeling through Stochastic Differential Equations},
  author={Song, Yang and Sohl-Dickstein, Jascha and Kingma, Diederik P and Kumar, Abhishek and Ermon, Stefano and Poole, Ben},
  booktitle={ICLR},
  year={2021}
}

@inproceedings{rombach2022latent,
  title={High-Resolution Image Synthesis with Latent Diffusion Models},
  author={Rombach, Robin and Blattmann, Andreas and Lorenz, Dominik and Esser, Patrick and Ommer, Björn},
  booktitle={CVPR},
  year={2022}
}

@inproceedings{saharia2022photorealistictexttoimagediffusionmodels,
      title={Photorealistic Text-to-Image Diffusion Models with Deep Language Understanding}, 
      author={Chitwan Saharia and William Chan and Saurabh Saxena and Lala Li and Jay Whang and Emily Denton and Seyed Kamyar Seyed Ghasemipour and Burcu Karagol Ayan and S. Sara Mahdavi and Rapha Gontijo Lopes and Tim Salimans and Jonathan Ho and David J Fleet and Mohammad Norouzi},
      booktitle={NIPS},
        year={2022} 
}

@inproceedings{nichol2021improved,
  title={Improved Denoising Diffusion Probabilistic Models},
  author={Nichol, Alex and Dhariwal, Prafulla},
  booktitle={ICML},
  year={2021}
}

@inproceedings{ho2022classifierfree,
  title={Classifier-Free Diffusion Guidance},
  author={Ho, Jonathan and Salimans, Tim},
  booktitle={NeurIPS Workshop},
  year={2021}
}

@inproceedings{zhang2023controlnet,
  title={Adding Conditional Control to Text-to-Image Diffusion Models},
  author={Zhang, Lvmin and Rao, Anyi and Agrawala, Maneesh},
  booktitle={ICLR},
  year={2023}
}

@inproceedings{zhang2024omnicontrol,
  title={OminiControl: Minimal and Universal Control for Diffusion Transformer},
  author={Zhenxiong Tan and Songhua Liu and Xingyi Yang and Qiaochu Xue and Xinchao Wang},
  booktitle={ICCV},
  year={2025}
}

@inproceedings{peebles2023dit,
  title={Scalable Diffusion Models with Transformers},
  author={Peebles, William and Xie, Saining},
  booktitle={ICCV},
  year={2023}
}

@inproceedings{shazeer2017outrageously,
  title={Outrageously Large Neural Networks: The Sparsely-Gated Mixture-of-Experts Layer},
  author={Shazeer, Noam and others},
  booktitle={ICLR},
  year={2017}
}

@inproceedings{lepikhin2020gshard,
  title={GShard: Scaling Giant Models with Conditional Computation and Automatic Sharding},
  author={Lepikhin, Dmitry and others},
  booktitle={ICLR},
  year={2021}
}

@inproceedings{fedus2022switch,
  title={Switch Transformers: Scaling to Trillion Parameter Models with Simple and Efficient Sparsity},
  author={Fedus, William and Zoph, Barret and Shazeer, Noam},
  booktitle={JMLR},
  year={2022},
}

@inproceedings{roller2021hash,
  title={Hash Layers for Large Sparse Models},
  author={Roller, Stephen and others},
  booktitle={NeurIPS},
  year={2021}
}

@inproceedings{park2019spade,
  title={Semantic Image Synthesis with SPADE},
  author={Park, Taesung and Liu, Ming-Yu and Wang, Ting-Chun and Zhu, Jun-Yan},
  booktitle={CVPR},
  year={2019}
}

@inproceedings{brock2019biggan,
  title={Large Scale GAN Training for High Fidelity Natural Image Synthesis},
  author={Brock, Andrew and Donahue, Jeff and Simonyan, Karen},
  booktitle={ICLR},
  year={2019}
}

@inproceedings{xu2022styleswin,
  title={StyleSwin: Transformer-based GAN for High-resolution Image Generation},
  author={Xu, Yang and others},
  booktitle={CVPR},
  year={2022}
}

@misc{liu2021fusedreamtrainingfreetexttoimagegeneration,
      title={FuseDream: Training-Free Text-to-Image Generation with Improved CLIP+GAN Space Optimization}, 
      author={Xingchao Liu and Chengyue Gong and Lemeng Wu and Shujian Zhang and Hao Su and Qiang Liu},
      year={2021},
      eprint={2112.01573},
      archivePrefix={arXiv},
      primaryClass={cs.CV},
      url={https://arxiv.org/abs/2112.01573}, 
}

@inproceedings{meng2022sdedit,
  title={SDEdit: Image Synthesis and Editing with Stochastic Differential Equations},
  author={Meng, Chenlin and Ho, Jonathan and Saharia, Chitwan and others},
  booktitle={ICLR},
  year={2022}
}

@inproceedings{hertz2022prompt2prompt,
  title={Prompt-to-Prompt Image Editing with Cross Attention Control},
  author={Hertz, Amir and Mokady, Ron and Tenenbaum, Jonathan and others},
  booktitle={ICLR},
  year={2023}
}

@inproceedings{mokady2023nulltext,
  title={Null-Text Inversion for Editing Real Images Using Guided Diffusion Models},
  author={Mokady, Ron and Hertz, Amir and Aberman, Kfir and others},
  booktitle={CVPR},
  year={2023}
}

@inproceedings{brooks2023instructpix2pix,
  title={InstructPix2Pix: Learning to Follow Image Editing Instructions},
  author={Brooks, Tim and Holynski, Aleksander and Efros, Alexei A.},
  booktitle={CVPR},
  year={2023}
}

@inproceedings{couairon2023diffedit,
  title={DiffEdit: Diffusion-Based Semantic Image Editing with Mask Guidance},
  author={Couairon, Guillaume and others},
  booktitle={ICLR},
  year={2023}
}

@inproceedings{caron2021emerging,
  title={Emerging Properties in Self-Supervised Vision Transformers},
  author={Caron, Mathilde and Touvron, Hugo and Misra, Ishan and others},
  booktitle={ICCV},
  year={2021}
}

@inproceedings{kingma2013auto,
  title={Auto-Encoding Variational Bayes},
  author={Kingma, Diederik P and Welling, Max},
  booktitle={ICLR},
  year={2014}
}

@inproceedings{hu2022lora,
  title={LoRA: Low-Rank Adaptation of Large Language Models},
  author={Hu, Edward and Shen, Yelong and Wallis, Phillip and others},
  booktitle={ICLR},
  year={2022}
}

@inproceedings{mou2023t2iadapterlearningadaptersdig,
      title={T2I-Adapter: Learning Adapters to Dig out More Controllable Ability for Text-to-Image Diffusion Models}, 
      author={Chong Mou and Xintao Wang and Liangbin Xie and Yanze Wu and Jian Zhang and Zhongang Qi and Ying Shan and Xiaohu Qie},
      booktitle={AAAI},
  year={2024}
}

@inproceedings{kervadec2019boundary,
  title     = {Boundary loss for highly unbalanced segmentation},
  author    = {Kervadec, Hoel and Bouchtiba, Jawad and Desrosiers, Christian and Granger, Eric and Dolz, Jose and Ayed, Ismail Ben},
  booktitle = {MIDL},
  year      = {2019}
}

@inproceedings{johnson2016perceptual,
  title     = {Perceptual losses for real-time style transfer and super-resolution},
  author    = {Johnson, Justin and Alahi, Alexandre and Fei-Fei, Li},
  booktitle = {ECCV},
  year      = {2016}
}

@inproceedings{sheynin2024emu,
  title     = {Emu-Edit: Precise Image Editing via Emu Diffusion Models},
  author={Shelly Sheynin and Adam Polyak and Uriel Singer and Yuval Kirstain and Amit Zohar and Oron Ashual and Devi Parikh and Yaniv Taigman},
  booktitle = {CVPR},
  year      = {2024}
}

@inproceedings{ace2024,
   title={ACE: All-round Creator and Editor Following Instructions via Diffusion Transformer}, 
      author={Ruipeng Wang and Junfeng Fang and Jiaqi Li and Hao Chen and Jie Shi and Kun Wang and Xiang Wang},
  booktitle = {ICLR},
  year      = {2026}
}

@misc{mao2025aceinstructionbasedimagecreation,
      title={ACE++: Instruction-Based Image Creation and Editing via Context-Aware Content Filling}, 
      author={Chaojie Mao and Jingfeng Zhang and Yulin Pan and Zeyinzi Jiang and Zhen Han and Yu Liu and Jingren Zhou},
      year={2025},
      eprint={2501.02487},
      archivePrefix={arXiv},
      primaryClass={cs.CV},
      url={https://arxiv.org/abs/2501.02487}, 
}

@inproceedings{omnigen2024,
  title={OmniGen: Unified Image Generation}, 
      author={Shitao Xiao and Yueze Wang and Junjie Zhou and Huaying Yuan and Xingrun Xing and Ruiran Yan and Chaofan Li and Shuting Wang and Tiejun Huang and Zheng Liu},
  booktitle = {CVPR},
  year      = {2025}
}

@misc{wu2025omnigen2,
      title={OmniGen2: Exploration to Advanced Multimodal Generation}, 
      author={Chenyuan Wu and Pengfei Zheng and Ruiran Yan and Shitao Xiao and Xin Luo and Yueze Wang and Wanli Li and Xiyan Jiang and Yexin Liu and Junjie Zhou and Ze Liu and Ziyi Xia and Chaofan Li and Haoge Deng and Jiahao Wang and Kun Luo and Bo Zhang and Defu Lian and Xinlong Wang and Zhongyuan Wang and Tiejun Huang and Zheng Liu},
      year={2025},
      eprint={2506.18871},
      archivePrefix={arXiv},
      primaryClass={cs.CV},
      url={https://arxiv.org/abs/2506.18871}, 
}

@inproceedings{mige2024,
  title     = {MIGE: Multi-Instruction Guided Editing via Diffusion Models},
  author={Tian, Xueyun and Li, Wei and Xu, Bingbing and Yuan, Yige and Wang, Yuanzhuo and Shen, Huawei},
  booktitle = {ACMMM},
  year      = {2025},
}

@inproceedings{anyedit,
      title={AnyEdit: Mastering Unified High-Quality Image Editing for Any Idea}, 
      author={Qifan Yu and Wei Chow and Zhongqi Yue and Kaihang Pan and Yang Wu and Xiaoyang Wan and Juncheng Li and Siliang Tang and Hanwang Zhang and Yueting Zhuang},
      booktitle = {CVPR},
        year = {2025},
}

@article{ma2025controllable,
  title={Controllable Video Generation: A Survey},
  author={Ma, Yue and Feng, Kunyu and Hu, Zhongyuan and Wang, Xinyu and Wang, Yucheng and Zheng, Mingzhe and He, Xuanhua and Zhu, Chenyang and Liu, Hongyu and He, Yingqing and others},
  journal={arXiv preprint arXiv:2507.16869},
  year={2025}
}

@inproceedings{ma2024followyouremoji,
  title={Follow-your-emoji: Fine-controllable and expressive freestyle portrait animation},
  author={Ma, Yue and Liu, Hongyu and Wang, Hongfa and Pan, Heng and He, Yingqing and Yuan, Junkun and Zeng, Ailing and Cai, Chengfei and Shum, Heung-Yeung and Liu, Wei and others},
  booktitle={SIGGRAPH Asia},
  year={2024}
}

@inproceedings{dreambooth,
  title     = {DreamBooth: Fine Tuning Text-to-Image Diffusion Models for Subject-Driven Generation},
  author    = {Ruiz, Nataniel and Li, Yuanzhen and Jampani, Varun and Pritch, Yael and Rubinstein, Michael and Aberman, Kfir},
  booktitle = {CVPR},
  year      = {2023}
}

@inproceedings{blipdiffusion,
  title     = {BLIP-Diffusion: Pre-trained Vision-Language Models for Zero-Shot Image-to-Image Translation},
  author    = {Li, Junnan and Li, Dongxu and Hu, Hexiang and Yang, Zhe and Yao, Kaixuan and Gao, Boyang and Wang, Yinfei and Amini, Lisa and Hoi, Steven C. H.},
  booktitle = {NIPS},
  year      = {2023}
}

@inproceedings{uno,
  title     = {UNO: Unified Neural Operator for Any-to-Any Generation},
  author    = {Wang, Yilun and Chen, Ziyi and Zhou, Hao and Tang, Wenqi and Lin, Zongxin},
  booktitle = {ICCV},
  year      = {2025}
}

@inproceedings{radford2021learning,
  title     = {Learning Transferable Visual Models From Natural Language Supervision},
  author    = {Radford, Alec and Kim, Jong Wook and Hallacy, Chris and Ramesh, Aditya and Goh, Gabriel and Agarwal, Sandhini and Sastry, Girish and Askell, Amanda and Mishkin, Pamela and Clark, Jack and Krueger, Gretchen and Sutskever, Ilya},
  booktitle = {ICML},
  year      = {2021},
}

@misc{oquab2023dinov2,
  title     = {DINOv2: Learning Robust Visual Features Without Supervision},
  author    = {Oquab, Maxime and Darcet, Tristan and Moutakanni, Theo and Vo, Huy and Szafraniec, Marc Szafraniec and Khalidov, Vasil and Fernandez, Pierre and Haziza, Daniel and Massa, Francisco and El-Nouby, Alaaeldin and Assran, Mahmoud and Ballas, Nicolas and Howes, Russell and Galuba, Wojciech and Bojanowski, Piotr and Neverova, Natalia and Vedaldi, Andrea and Rabbat, Mike and LeCun, Yann and Caron, Mathilde},
  journal   = {arXiv preprint arXiv:2304.07193},
  year      = {2023},
  url       = {https://arxiv.org/abs/2304.07193},
}

@inproceedings{zhang2024magicbrushmanuallyannotateddataset,
      title={MagicBrush: A Manually Annotated Dataset for Instruction-Guided Image Editing}, 
      author={Kai Zhang and Lingbo Mo and Wenhu Chen and Huan Sun and Yu Su},
      booktitle = {NIPS},
        year      = {2023},
}

@misc{flux2024,
    author={Black Forest Labs},
    title={FLUX},
    year={2024},
    howpublished={\url{https://github.com/black-forest-labs/flux}},
}

@inproceedings{peng2025dreambenchhumanalignedbenchmarkpersonalized,
      title={DreamBench++: A Human-Aligned Benchmark for Personalized Image Generation}, 
      author={Yuang Peng and Yuxin Cui and Haomiao Tang and Zekun Qi and Runpei Dong and Jing Bai and Chunrui Han and Zheng Ge and Xiangyu Zhang and Shu-Tao Xia},
      year={2025},
      booktitle = {ICLR},
}

@misc{wei2025omnieditbuildingimageediting,
      title={OmniEdit: Building Image Editing Generalist Models Through Specialist Supervision}, 
      author={Cong Wei and Zheyang Xiong and Weiming Ren and Xinrun Du and Ge Zhang and Wenhu Chen},
      year={2025},
      eprint={2411.07199},
      archivePrefix={arXiv},
      primaryClass={cs.CV},
      url={https://arxiv.org/abs/2411.07199}, 
}

@inproceedings{zhang2025icedit,
  title     = {In-Context Edit: Enabling Instructional Image Editing with In-Context Generation in Large-Scale Diffusion Transformers},
  author    = {Zhang, Zechuan and Xie, Ji and Lu, Yu and Yang, Zongxin and Yang, Yi},
  booktitle = {NIPS},
  year      = {2025}
}

@misc{wang2025moditlearninghighlyconsistent,
      title={MoDiT: Learning Highly Consistent 3D Motion Coefficients with Diffusion Transformer for Talking Head Generation}, 
      author={Yucheng Wang and Dan Xu},
      year={2025},
      eprint={2507.05092},
      archivePrefix={arXiv},
      primaryClass={cs.CV},
      url={https://arxiv.org/abs/2507.05092}, 
}

@misc{hui2024hqedithighqualitydatasetinstructionbased,
      title={HQ-Edit: A High-Quality Dataset for Instruction-based Image Editing}, 
      author={Mude Hui and Siwei Yang and Bingchen Zhao and Yichun Shi and Heng Wang and Peng Wang and Yuyin Zhou and Cihang Xie},
      year={2024},
      eprint={2404.09990},
      archivePrefix={arXiv},
      primaryClass={cs.CV},
      url={https://arxiv.org/abs/2404.09990}, 
}

@inproceedings{wang2025rep,
  title={Rep-MTL: Unleashing the Power of Representation-level Task Saliency for Multi-Task Learning},
  author={Wang, Zedong and Li, Siyuan and Xu, Dan},
  booktitle={ICCV},
  year={2025}
}

@inproceedings{ultraedit,
      title={UltraEdit: Instruction-based Fine-Grained Image Editing at Scale}, 
      author={Haozhe Zhao and Xiaojian Ma and Liang Chen and Shuzheng Si and Rujie Wu and Kaikai An and Peiyu Yu and Minjia Zhang and Qing Li and Baobao Chang},
      year={2024},
      booktitle = {NIPS}, 
}

@misc{chen2024zeroshotimageeditingreference,
      title={Zero-shot Image Editing with Reference Imitation}, 
      author={Xi Chen and Yutong Feng and Mengting Chen and Yiyang Wang and Shilong Zhang and Yu Liu and Yujun Shen and Hengshuang Zhao},
      year={2024},
      eprint={2406.07547},
      archivePrefix={arXiv},
      primaryClass={cs.CV},
      url={https://arxiv.org/abs/2406.07547}, 
}

@inproceedings{chen2024anydoorzeroshotobjectlevelimage,
      title={AnyDoor: Zero-shot Object-level Image Customization}, 
      author={Xi Chen and Lianghua Huang and Yu Liu and Yujun Shen and Deli Zhao and Hengshuang Zhao},
      year={2024},
      booktitle = {CVPR}, 
}

@inproceedings{sun2025ecditscalingdiffusiontransformers,
      title={EC-DIT: Scaling Diffusion Transformers with Adaptive Expert-Choice Routing}, 
      author={Haotian Sun and Tao Lei and Bowen Zhang and Yanghao Li and Haoshuo Huang and Ruoming Pang and Bo Dai and Nan Du},
      year={2025},
      booktitle = {ICLR}, 
}

@inproceedings{meng2024instructgiegeneralizableimageediting,
      title={InstructGIE: Towards Generalizable Image Editing}, 
      author={Zichong Meng and Changdi Yang and Jun Liu and Hao Tang and Pu Zhao and Yanzhi Wang},
      year={2024},
      booktitle = {ECCV},
}

@misc{xu2025incontextbrushzeroshotcustomized,
      title={In-Context Brush: Zero-shot Customized Subject Insertion with Context-Aware Latent Space Manipulation}, 
      author={Yu Xu and Fan Tang and You Wu and Lin Gao and Oliver Deussen and Hongbin Yan and Jintao Li and Juan Cao and Tong-Yee Lee},
      year={2025},
      eprint={2505.20271},
      archivePrefix={arXiv},
      primaryClass={cs.CV},
      url={https://arxiv.org/abs/2505.20271}, 
}

@inproceedings{tumanyan2022plugandplaydiffusionfeaturestextdriven,
      title={Plug-and-Play Diffusion Features for Text-Driven Image-to-Image Translation}, 
      author={Narek Tumanyan and Michal Geyer and Shai Bagon and Tali Dekel},
      year={2023},
      booktitle = {CVPR},
}

@misc{wang2024instantidzeroshotidentitypreservinggeneration,
      title={InstantID: Zero-shot Identity-Preserving Generation in Seconds}, 
      author={Qixun Wang and Xu Bai and Haofan Wang and Zekui Qin and Anthony Chen and Huaxia Li and Xu Tang and Yao Hu},
      year={2024},
      eprint={2401.07519},
      archivePrefix={arXiv},
      primaryClass={cs.CV},
      url={https://arxiv.org/abs/2401.07519}, 
}
}

\clearpage
\maketitlesupplementary

\appendix


This appendix provides complete supplementary material to the main manuscript and is organized as follows:

\begin{itemize}
    \item \textbf{Appendix~\ref{app:dataset_details}: Dataset and Implementation Details.} We detail the construction of training dataset in Appendix~\ref{app:dataset}, including the targeted design of mask-aware image pairs and the generation pipeline. We also provide a systematic comparison with existing public datasets.
Appendix~\ref{app:impl_arch_train} presents implementation specifications, including network architectures, optimization protocols, training schedules, and hyperparameter configurations in our experiments.
    \item \textbf{Appendix~\ref{app:qualitative}: Extended Qualitative Comparisons.} We report additional qualitative comparisons for instruction-based (Appendix~\ref{app:instruction}) and subject-driven (Appendix~\ref{app:subject}) editing that could not be included in main text due to space limitations, encompassing mainstream tasks such as object removal, replacement, and style transfer. For each task, we include detailed per-category and per-edit-type samples and further discuss the behavior and capabilities of CARE-Edit. We provide a 
    \href{https://care-edit.github.io/}{project page}, where we host more qualitative examples and interactive visualizations.

    \item \textbf{Appendix~\ref{app:analysis}: Additional Empirical Analysis.} We present the empirical analysis and visualizations of latent attention maps in Appendix~\ref{app:analysis} to validate the efficacy of specialized experts. Figure~\ref{fig:expert_full} demonstrates how each expert in CARE-Edit, including condition-aware routing, reference-guided subject preservation, and mask-aware control, contributes to effective task-specific learning over diffusion timesteps.
\end{itemize}

\section{Dataset and Implementation Details}
\label{app:dataset_details}

\subsection{Training Dataset}
\label{app:dataset}

\paragraph{Base corpora.}
To equip CARE-Edit with diverse editing capabilities, we collect data from several high-quality sources, targeting four key editing tasks: \emph{instruction-based editing}, \emph{object removal/replacement}, and \emph{style transfer}.
(i) Instruction-based edits are drawn from MagicBrush~\citep{zhang2024magicbrushmanuallyannotateddataset} and OmniEdit~\cite{wei2025omnieditbuildingimageediting}, which provide rich natural-language instructions paired with real-world images. (ii) and (iii) Removal and replacement samples are sourced from UNO~\cite{uno} subset.
(iv) Style transfer data is enriched using AnyEdit~\citep{anyedit}, which contains both the fine-grained appearance- and style-level instructions (\eg, ``convert to watercolor painting'').

\begin{figure}[t]
\centering
\includegraphics[width=\linewidth]{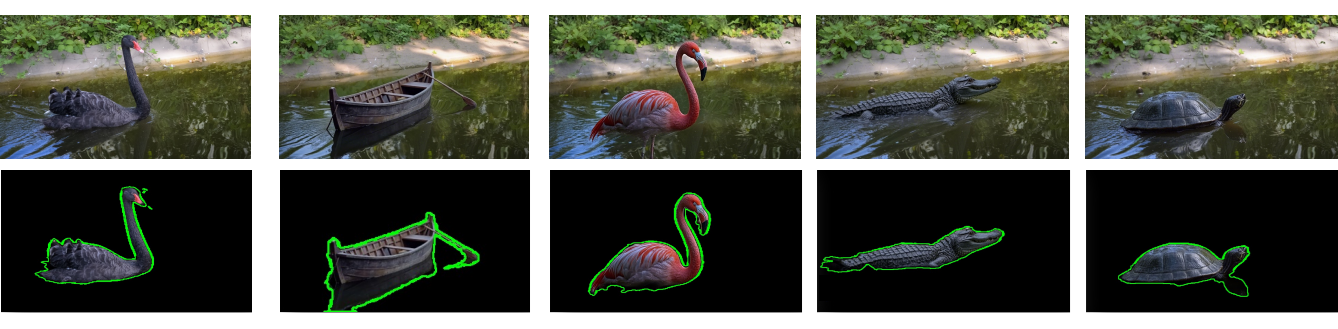}
\vspace{-1.70em}
\caption{\textbf{Pipeline for Mask-aware Image-Pair Generation.} We use a GPT-Image-1 and VLM-based pipeline to create our training data.
  Starting with a reference subject on a white background, we generate diverse scene descriptions and corresponding images using an image-to-image model.
  This yields high-quality image pairs with consistent backgrounds but varying foregrounds, annotated with precise segmentation masks ($\mathcal{M}$) and bounding boxes ($\mathcal{B}$).}
\vspace{-1.00em}
\label{fig:mask_pairs}
\end{figure}

\vspace{-0.40em}
\paragraph{Motivation for Subjects200K.}
A critical limitation of purely instruction-based datasets is the spatial underspecification of edits. Models are often forced to infer the edit location solely from language, leading to ambiguity.
To address this, we require a dataset with precise object masks to explicitly teach the model to align edits with spatial constraints. We thus select Subjects200K~\citep{zhang2024omnicontrol} as our foundation for two reasons:
(i) it offers high-quality foreground masks for a diverse taxonomy of objects and humans; and
(ii) reference images are captured on clean white backgrounds, which simplifies downstream composition and in-context editing.
From this source, we construct a $20\mathrm{K}$ subset where each sample comprises an image, a fine-grained mask, an instruction, and an optional reference image, enabling CARE-Edit to learn region-specific editing while keeping subject identity.

\vspace{-0.40em}
\paragraph{Mask-Aware Image-Pair Generation.}
As shown in Figure~\ref{fig:mask_pairs}, we construct background-consistent image pairs with varying foregrounds using a GPT-based generation pipeline. Starting with a Subjects200K~\citep{zhang2024omnicontrol} reference subject, we (i) sample a scene template and a set of subjects, and (ii) query the GPT-image-1 model to synthesize images that share a consistent background but feature diverse foreground objects. (iii) Extract a high-resolution fine mask $\mathcal{M} \in \{0,1\}^{H \times W}$ for foreground using an off-the-shelf segmentation model, followed by manual filtering to ensure quality. In practice, we instantiate these templates with category names (\eg, \textit{swan}, \textit{boat}, \textit{flamingo}) and short descriptions of target background (\eg, \textit{“floating on a calm river near the shore”}). To improve robustness against imperfect user inputs at inference time, we also derive a coarse mask $\mathcal{B}$ in training, defined as the tight axis-aligned bounding box of the fine mask $\mathcal{M}$:
\begin{align}
\Omega &= \{(x,y)\mid \mathcal{M}(x,y) = 1\}, \\
x_{\min} &= \min_{(x,y)\in\Omega} x,\quad
x_{\max} = \max_{(x,y)\in\Omega} x, \\
y_{\min} &= \min_{(x,y)\in\Omega} y,\quad
y_{\max} = \max_{(x,y)\in\Omega} y, \\
\mathcal{B}(x,y) &=
\begin{cases}
1, & x_{\min} \le x \le x_{\max} \land y_{\min} \le y \le y_{\max}, \\
0, & \text{otherwise},
\end{cases}
\end{align}
where $x_{\min}, x_{\max}, y_{\min}, y_{\max}$ are the extrema of coordinates in $\Omega$. We utilize $\mathcal{M}$ for the pixel-accurate supervision (\eg, boundary consistency loss) and $\mathcal{B}$ as a coarse spatial prior for the condition-aware expert routing in CARE-Edit.

\begin{figure*}[t]
    \centering
    \includegraphics[width=\linewidth]{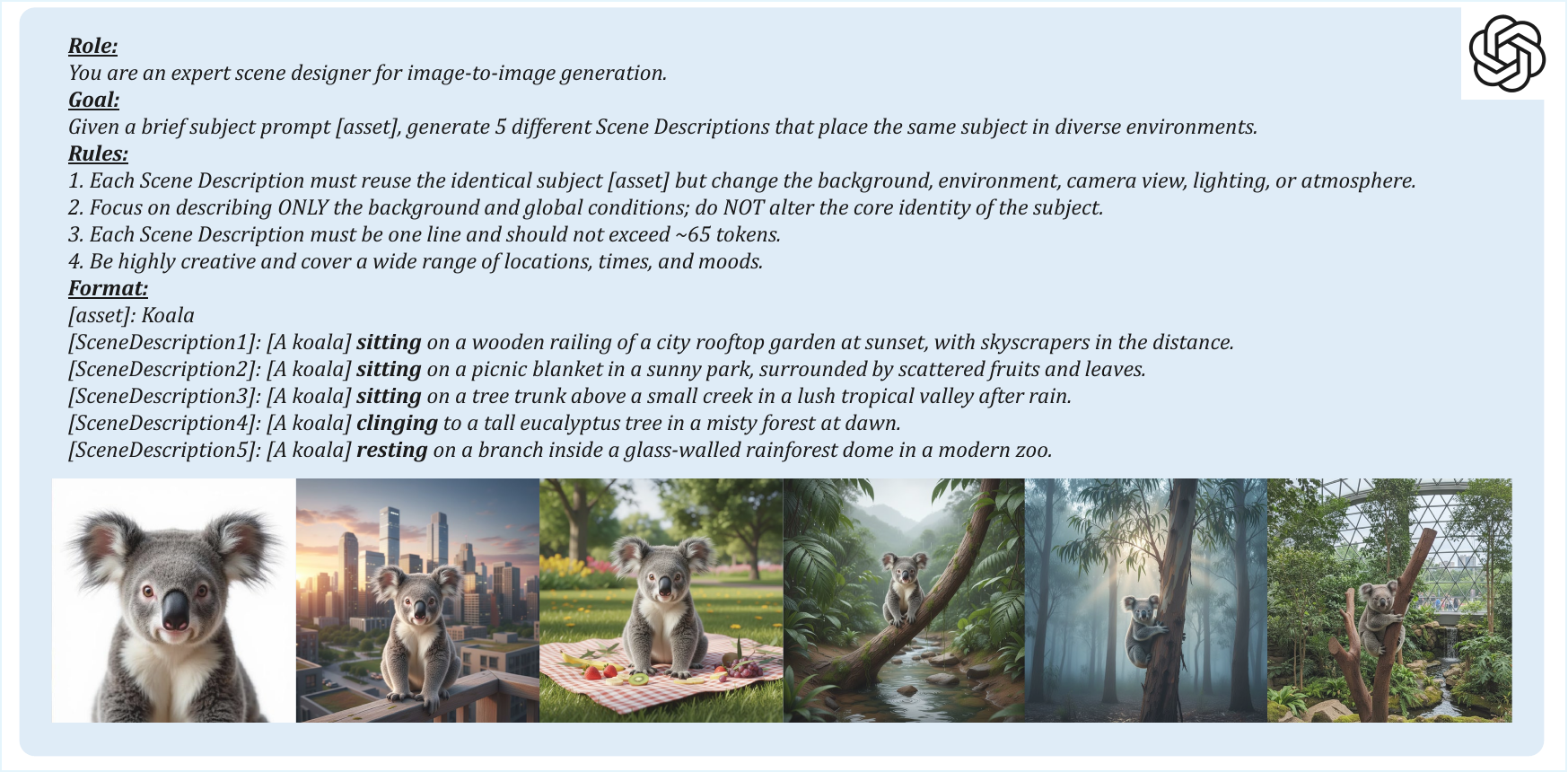}
    \caption{\textbf{GPT Prompting for Scene Diversity.} We explicitly instruct the LLM to generate varied scene descriptions (\eg, ``sunny park'') while keeping the subject (\eg, ``Koala'') constant. As such, by synthesizing multiple environments for the identical subject entity, we create training data that compels the image editing model to learn robust subject grounding independent of the background correlations.}
    \label{fig:gpt1}
\end{figure*}

\vspace{-0.40em}
\paragraph{Prompt Taxonomy.}
To systematically construct diverse yet controllable training triplets~\cite{ma2025controllable, ma2024followyouremoji} for our CARE-Edit, we organize
the generation pipeline along two axes:

\textbf{(1) Category and Operation.}
We factorize the synthetic space into (i) \emph{Categories} (people, animals, everyday objects, stylized assets) and (ii) \emph{Operation Types} (instruction-based, removal, replacement, style transfer, multi-subject cases).

\textbf{(2) Scene-level Templates.}
Conditioned on a subject category, we query the LLM to
generate multiple scene descriptions. Crucially, the scene prompt is constrained to a single line and constrains the generation to modify only the environment, lighting, camera view, or overall atmosphere, while strictly preserving the core subject identity or attributes. For each \emph{Category}, we sample five such scene descriptions.
This separation of subject and scene allows us to reuse identical subjects across
heterogeneous contexts, and in turn to build foreground-consistent but
background-varying pairs for mask-aware training. Please refer to Figure~\ref{fig:gpt1} for details.

\begin{figure*}[t]
    \centering
    \includegraphics[width=\linewidth]{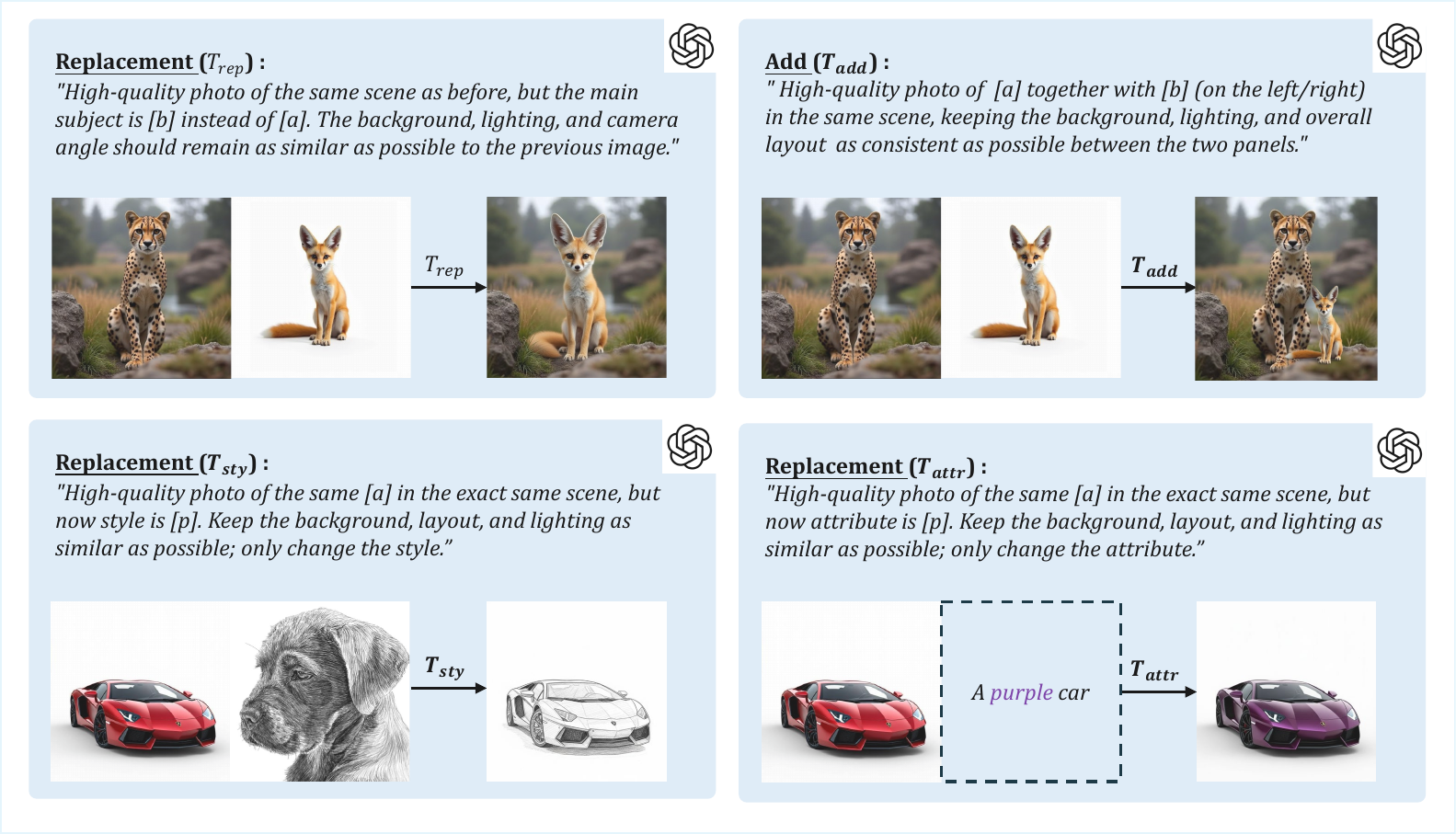}
    \vspace{-1.60em}
    \caption{\textbf{Task-Specific GPT Prompt Templates.} We visualize the templates used to construct training triplets for: (i) \textit{Replacement} (Top-Left); (ii) \textit{Addition} (Top-Right); and (iii) \textit{Style/Attribute Change} (Bottom). By constraining the LLM to modify specific slots (\eg, subject identity) while holding scene descriptions constant, it ensures the resulting image pairs possess consistent backgrounds.}
    \label{fig:gpt2}
\end{figure*}

\textbf{(3) Task-level Templates.}
Given two subject descriptions $a$ and $b$, a scene
description $s$, and a style phrase $p$ (\eg, ``in a retro vintage
style''), we instantiate three mask-friendly templates:
(i) \emph{Replacement} ($T_{\text{rep}}$) keeps the background unchanged while
swapping the main subject from $a$ to $b$;
(ii) \emph{Addition} ($T_{\text{add}}$) forms a diptych where one panel contains
only $a$ and the other contains both $a$ and $b$ under the same scene $s$;
(iii) \emph{style / attribute change} ($T_{\text{sty}}$ / $T_{\text{attr}}$) preserves the subject
identity and scene but alters only high-level appearance attributes $p$; All templates explicitly ask the generator to keep background layout, lighting,
and camera viewpoint as similar as possible across paired images, so that the
resulting pairs differ primarily in well-localized foreground regions.
This naturally aligns with our fine masks $\mathcal{M}$ and coarse boxes
$\mathcal{B}$, and provides clean supervision for subject-centric, mask-aware
editing. Please view Figure~\ref{fig:gpt2} for details.

\vspace{-0.40em}
\paragraph{Data Efficiency vs Previous Training Setups.}
Despite being trained on a significantly smaller data corpus ($\sim 120\mathrm{K}$ triplets in total) compared to recent state-of-the-art editing and personalization baselines, CARE-Edit achieves superior performance. Table~\ref{tab:dreambenchpp_half_A} reports multiple-object results on DreamBench++~\citep{peng2025dreambenchhumanalignedbenchmarkpersonalized}, together with the approximate scale of the training data used by each method.
The results show that CARE-Edit outperforms strong baselines such as OmniControl~\citep{zhang2024omnicontrol}, UNO~\citep{uno}, and OmniGen2~\citep{wu2025omnigen2} on all multiple-object metrics, despite relying on substantially fewer training samples.
This suggests that our mask-aware, subject-centric curriculum applied in CARE-Edit and the curated multi-paired construction are substantially more data-efficient than simply scaling up instruction-only datasets, particularly for applications with complicated multi-object compositions.

\begin{table}[t]
\centering
\caption{
Quantitative results on the \emph{multiple-object} subset of DreamBench++~\citep{peng2025dreambenchhumanalignedbenchmarkpersonalized}.
We report three metrics along with the approximate number of training examples used by each method.
The best and second-best results are marked in \textbf{bold} and \underline{underlined}, respectively.
}
\label{tab:dreambenchpp_half_A}
\vspace{-0.40em}
\resizebox{0.40\textwidth}{!}{
\begin{tabular}{l r *{3}{c}}
\toprule[1.2pt]
Method & \#Train data & DINO‑I $\uparrow$ & CLIP‑I $\uparrow$ & CLIP‑T $\uparrow$ \\
\midrule
OmniControl~\citep{zhang2024omnicontrol}
  & \phantom{0}\textasciitilde$1M$ & 0.501 & 0.641 & 0.316 \\
UNO~\citep{uno}
  & \phantom{0}\textasciitilde$1M$ & 0.508 & 0.649 & 0.303 \\
OmniGen2~\citep{wu2025omnigen2}
  & $\geq533K$ & \underline{0.560} & \underline{0.713} & \underline{0.319} \\
\midrule
\textbf{CARE‑Edit (Ours)}
  & $120K$         & \textbf{0.568} & \textbf{0.720} & \textbf{0.327} \\
\bottomrule[1.2pt]
\end{tabular}}
\vspace{-0.30em}
\end{table}

\subsection{Implementation Details}
\label{app:impl_arch_train}

\paragraph{Backbone and Training Setup.}
CARE-Edit is built upon the FLUX.1~\cite{flux2024} variant of the Rectified Flow Transformer family. Unless otherwise specified, we select \texttt{FLUX.1-dev} as the backbone for all experiments, as it offers a good balance between visual quality and training stability in the editing setup. Following the design of OmniControl~\cite{zhang2024omnicontrol}, we employ condition-aware modules via LoRA~\cite{hu2022lora} on top of the base model, and keep the original backbone weights frozen during fine-tuning. We adopt a standard LoRA rank~\cite{hu2022lora} of $r=4$ for all attention modules, and only enable the LoRA branches~\cite{hu2022lora} when processing condition-related tokens. For regular text-only tokens, the LoRA scale is set to zero so that the backbone behaves identically to the original FLUX.1~\cite{flux2024} model. All models are trained on 8$\times$NVIDIA L20 GPUs, which corresponds to roughly 800 GPU hours in total. We use a per-GPU batch size of 1 with gradient accumulation over 8 steps (effective batch size of 8). For most experiments, we follow a two-stage training schedule: the model is first trained for $40\mathrm{K}$ iterations on basic, single-subject samples, and then switches to complex multi-subject data for the remaining $60\mathrm{K}$ iterations. We also apply EMA to the LoRA parameters~\cite{hu2022lora} with a decay rate of $0.999$.

\vspace{-0.40em}
\paragraph{Loss Functions and Hyperparameters.}
The total training objective $\mathcal{L}_{\text{CARE}}$ combines the standard diffusion reconstruction loss $\mathcal{L}_{\text{diff}}$ (Sec.~\ref{sec:preliminaries}) with three regularization terms as:

\begin{equation}
\mathcal{L}_{\text{CARE}}
=
\mathcal{L}_{\text{diff}}
+ \lambda_{\text{load}}\,\mathcal{L}_{\text{load}}
+ \lambda_{\text{mask}}\,\mathcal{L}_{\text{mask}}
+ \lambda_{\text{mix}}\,\mathcal{L}_{\text{mix}}
\label{eq:lcare_A}
\end{equation}
where $\mathcal{L}_{\text{load}}$ ensures balanced expert utilization (Sec.~\ref{sec:method_overview}), $\mathcal{L}_{\text{mask}}$ enforces the boundary consistency (Sec.~\ref{sec:mask_repaint}), and $\mathcal{L}_{\text{mix}}$ encourages spatial smoothness in the mixture map (Sec.~\ref{sec:latent_mixture}). To prioritize the reconstruction term while maintaining regularization, we empirically set small weights to regularizers:
\begin{equation}
(\lambda_{\text{load}},\,\lambda_{\text{mask}},\,\lambda_{\text{mix}})
= (0.01,\, 0.1,\, 0.05).
\end{equation}
where the hyperparameters were identified according to prior works~\citep{sun2025ecditscalingdiffusiontransformers, wang2025rep} and fixed for all experiments w/o extra tuning.

\vspace{-0.40em}
\paragraph{Evaluation.}
We evaluate CARE-Edit on four representative image editing tasks across three diverse benchmarks that probe different aspects of contextual image editing:
\begin{itemize}
    \item \textbf{EMU-Edit}~\cite{sheynin2024emu}: this benchmark tests both the object-level and attribute-level modifications on real-world photos using fine-grained text prompt descriptions.
    \item \textbf{MagicBrush}~\cite{zhang2024magicbrushmanuallyannotateddataset}: this dataset involves complex, region-based editing tasks guided by free-form natural language instructions and the user-provided masks.
    \item \textbf{DreamBench++}~\cite{peng2025dreambenchhumanalignedbenchmarkpersonalized}: this benchmark evaluates personalized subject-driven image editing and composition, covering single-object and complex multiple-object scenarios.
\end{itemize}

We follow the official data splits and evaluation subsets for each benchmark whenever available. Following the data processing pipelines in OmniControl~\cite{zhang2024omnicontrol} and UNO~\cite{uno}, we resize images to $512{\times}512$ while preserving aspect ratio.

\section{Extended Qualitative Comparisons}
\label{app:qualitative}

Appendix~\ref{app:qualitative} reports additional experimental results that could not be included in the main paper due to space limitations. In particular, we organize these results by task to demonstrate the model's robustness in handling diverse semantic demands, from subtle attribute changes to complex scene re-contextualization. For each task, we include representative per-category and per-edit-type samples and briefly discuss the behavior of CARE-Edit across different settings. 

\subsection{Instruction-based Image Editing}
\label{app:instruction}

\begin{figure*}[t]
    \centering
    \includegraphics[width=\textwidth]{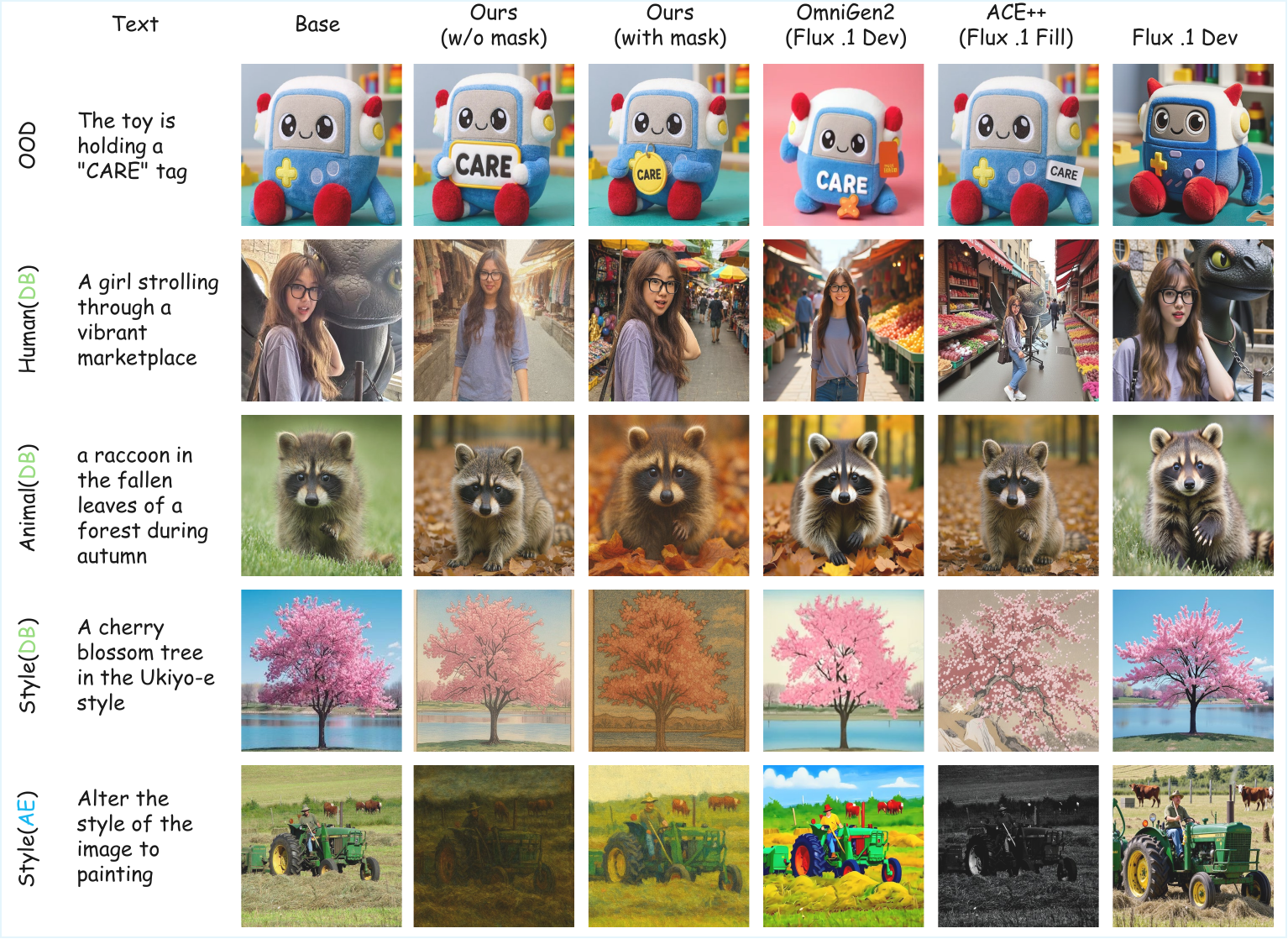}
    \vspace{-1.00em}
    \caption{\textbf{Instruction-based Editing on EMU-Edit and MagicBrush.} We compare CARE-Edit against existing unified editors~\citep{wu2025omnigen2, ace2024}. Rows exhibit challenging scenarios such as text rendering (``CARE tag''), global style transfer (``Ukiyo-e''), and complex object insertions. \textbf{(1) Text and Geometric Fidelity (Row 1):} The geometric rigidity of the tag and the legibility of the text are paramount. While ACE++ and OmniGen2 correctly interpret the semantic intent, they suffer from \textit{structural drift}, resulting in warped tag boundaries and deformed glyphs. CARE-Edit, particularly the masked variant, utilizes the Mask Expert to enforce spatial constraints, producing orthogonal tag geometry and crisp, readable text. \textbf{(2) Identity Preservation (Row 2):} CARE-Edit preserves the subject's facial identity and hair texture significantly better than baselines, which tend to over-blend the subject into the crowd (identity dilution) or generate a generic face.
    \textbf{(3) Style Disentanglement (Row 4 \& 5):} In the \textit{``Ukiyo-e''} and \textit{``Painting''} style transfer tasks, baselines often hallucinate new objects or flatten the entire scene into a texture map. CARE-Edit disentangles style from structure. The \textit{Base Expert} maintains the complex branching of the cherry blossom tree and the mechanical details of the tractor, while \textit{Reference Expert} strictly applies the artistic texture to the environments.}
    \label{fig:A1}
\end{figure*}

In this subsection, we introduce and discuss qualitative results on instruction-based image editing on EMU-Edit~\cite{sheynin2024emu} and MagicBrush~\cite{zhang2024magicbrushmanuallyannotateddataset}. Figure~\ref{fig:A1} shows a large-scale visual comparison between our method and several strong baselines, including OmniGen2~\cite{wu2025omnigen2}, ACE++~\cite{mao2025aceinstructionbasedimagecreation}, and the vanilla FLUX.1-Dev~\cite{flux2024} backbone. The examples cover typical instruction-based edits such as style changes, attribute modifications, and cases involving visible text (\eg, a toy holding a ``CARE'' label).

Qualitatively, CARE-Edit (especially the masked variant) follows the textual instructions while better preserving unedited content and fine-grained structures. Compared to SOTA methods, our method produces fewer spurious background changes and sharper, more localized boundaries at the edited regions. These trends are consistent with the quantitative gains reported in the main paper.

\subsection{Subject-driven Contextual Image Editing}
\label{app:subject}

In this subsection, we provide more results on subject-driven contextual editing, primarily on DreamBench++~\cite{peng2025dreambenchhumanalignedbenchmarkpersonalized}. The goal is not only to preserve subject identity (\eg, a particular person, pet, or product) but also to compose the subject into new contexts with complex surroundings and interactions.

A key motivation behind our design is that, for this class of tasks, it is often difficult to resolve the \emph{relative size and placement} of the reference objects in the base image using text prompts and the backbone model alone. Instructions such as “The man is holding a camera.” or “Add a Rubik’s Cube next to the sneaker.” do not uniquely determine how large the inserted object should be or where it should appear. To address this, CARE-Edit incorporates a user-provided mask as an additional control signal. Even when the mask is coarse, it specifies the intended location and approximate size of the edited content, disambiguating the spatial relationship between the reference objects and the base subject.

\begin{figure*}[h]
    \centering
    \includegraphics[width=0.85\textwidth]{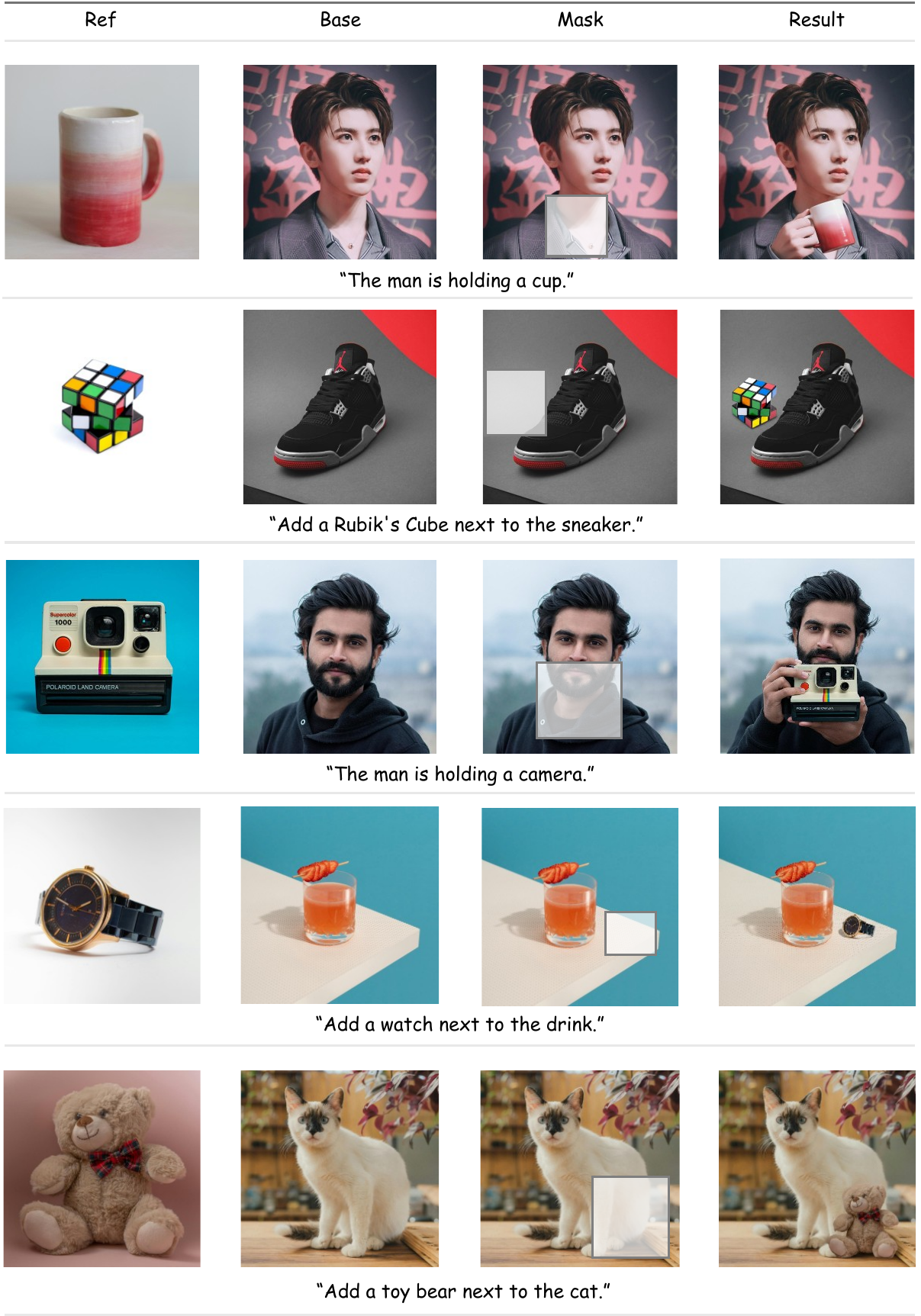}
    \vspace{-0.3em}
    \caption{\textbf{Complex Contextual Editing Results.} Multi-condition examples requiring harmonization of subject identity, mask constraints, and text prompts.
  CARE-Edit successfully composes subjects into disparate environments while respecting the user-provided spatial layout.}
    \label{fig:B1}
\end{figure*}

Figure~\ref{fig:B1} illustrates this design with representative cases such as “The man is holding a cup.”,  “Add a watch next to the drink.”, and “Add a toy bear next to the cat.”. In all these examples, CARE-Edit produces edits where the inserted objects have plausible geometry and scale, while the main subject’s identity, pose, and global lighting are preserved. This mask-guided formulation enables reliable subject-driven contextual editing in scenarios such as personalized product shots and multi-object layout design, where precise control over relative size and placement is crucial.

\subsection{More Results on Diverse Editing Tasks}
\label{app:tasks}

\begin{figure*}[t]
    \centering
    \includegraphics[width=\textwidth]{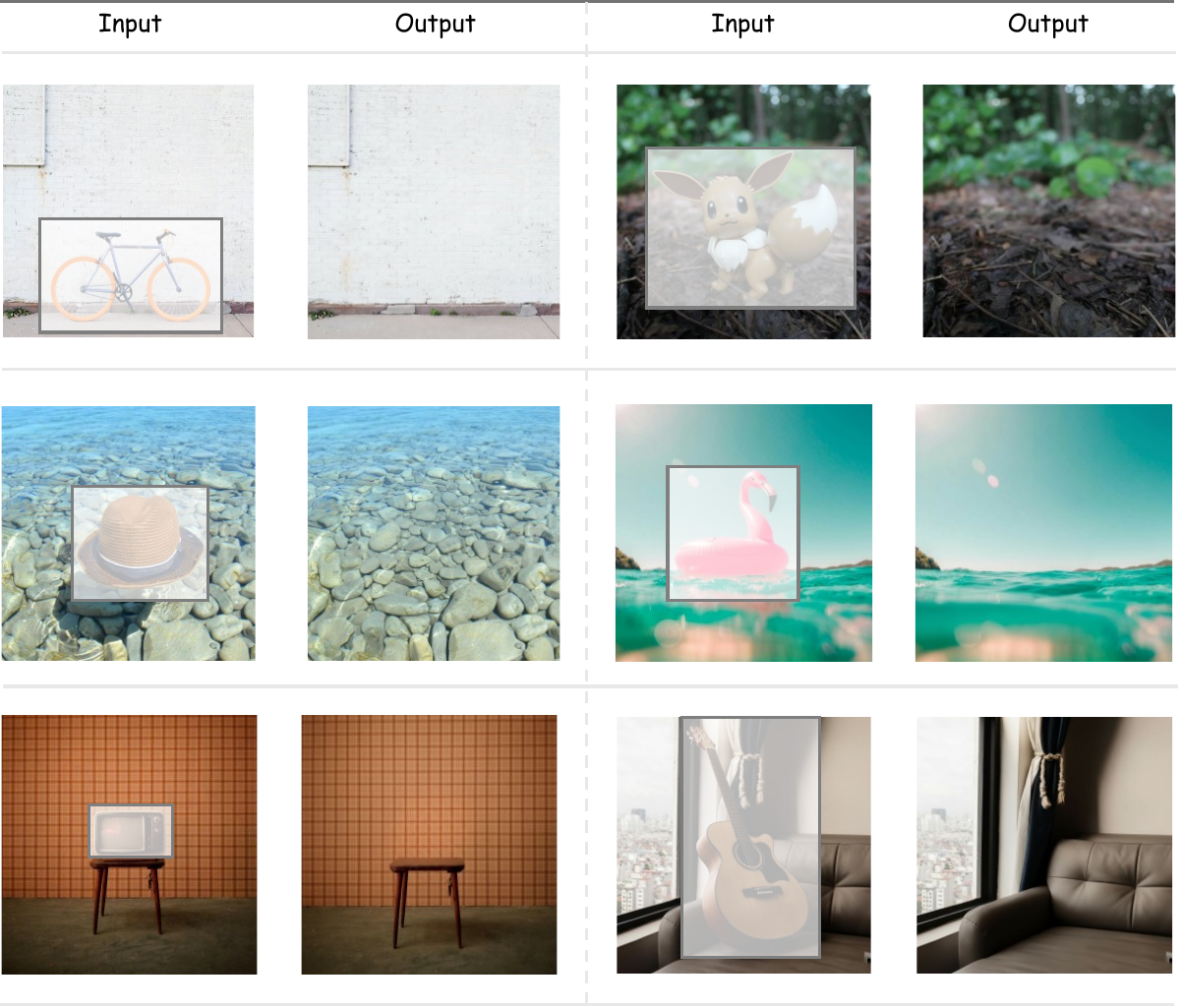}
    \vspace{-1.20em}
    \caption{\textbf{Object Removal Results.} This task requires the model to remove the foreground object and hallucinate a plausible background (\eg, sofa fabric, wall patterns) that is consistent with surrounding scene context. \textbf{(i) Top and Middle:} CARE-Edit successfully handles stochastic textures, such as natural water ripples and uneven stone surfaces, filling the void with spatially coherent content.
    \textbf{(ii) Bottom-Left:} A stress test for structural consistency. Removing the television requires reconstructing the rigid grid pattern of the wallpaper. CARE-Edit accurately hallucinates the missing tiles, maintaining the correct perspective and alignment without the blurring or geometric warping.}
    \label{fig:A3}
\end{figure*}

\begin{figure*}[t]
    \centering
    \includegraphics[width=\textwidth]{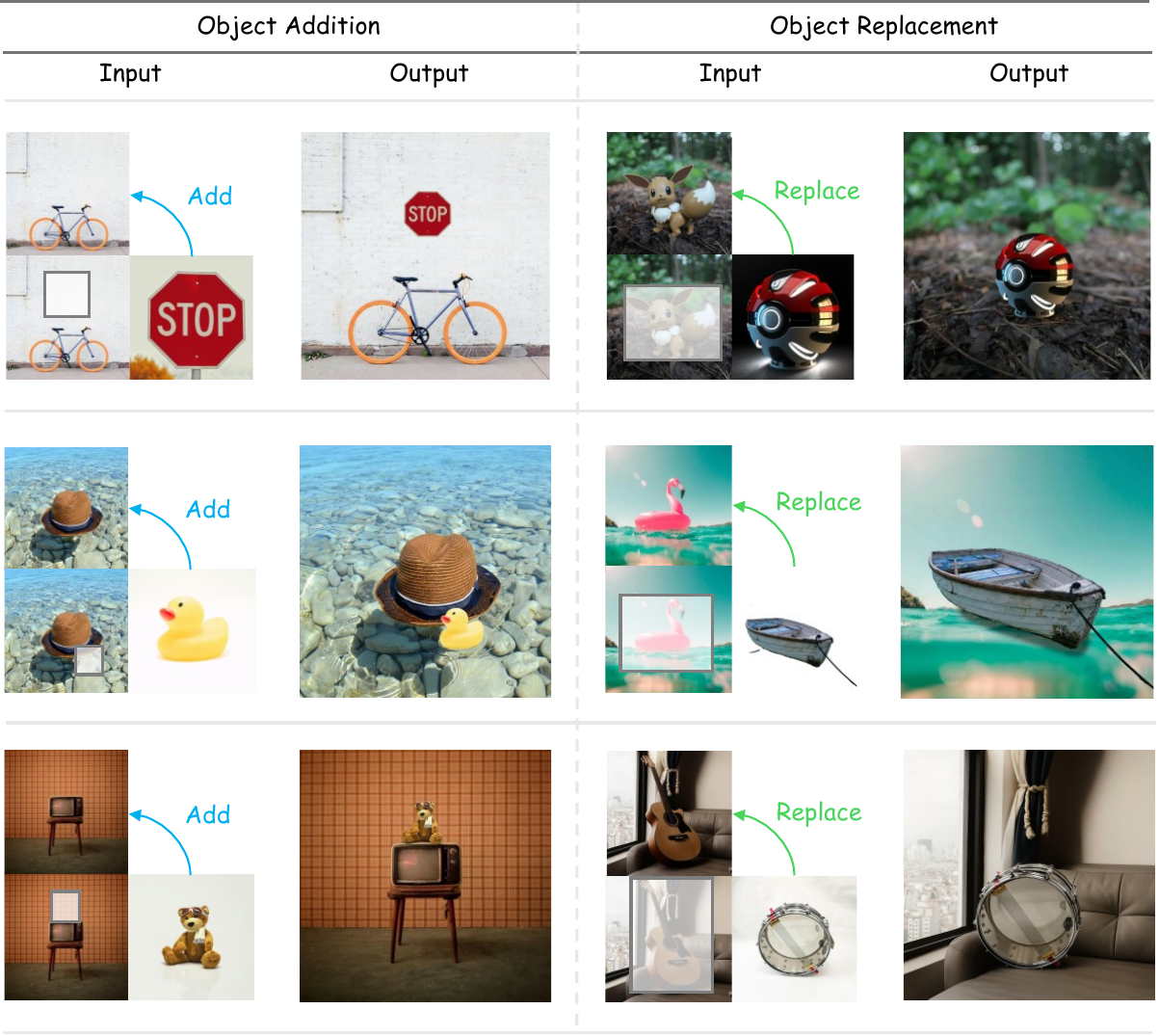}
    \vspace{-1.20em}
    \caption{\textbf{Object Addition and Replacement Results.}
    CARE-Edit demonstrates precise control over scene composition, inserting or swapping objects while rigorously adhering to environmental constraints. 
    \textbf{(i) Object Addition (Left):} CARE-Edit introduces new elements that respect physical laws. Note that how the added rubber duck (middle) is generated with accurate water reflections and surface interaction.
    \textbf{(ii) Object Replacement (Right):} CARE-Edit handles drastic changes in structure and material while maintaining lighting consistency. In the top-right example, replacing a furry pokemon creature with a pokemon ball (metallic sphere), CARE-Edit correctly renders specular highlights and casts realistic shadows onto the complex dirt terrain, ensuring the newly added object sits naturally within the depth of field.}
    \label{fig:A4}
\end{figure*}

\begin{figure*}[t]
    \centering
    \includegraphics[width=\textwidth]{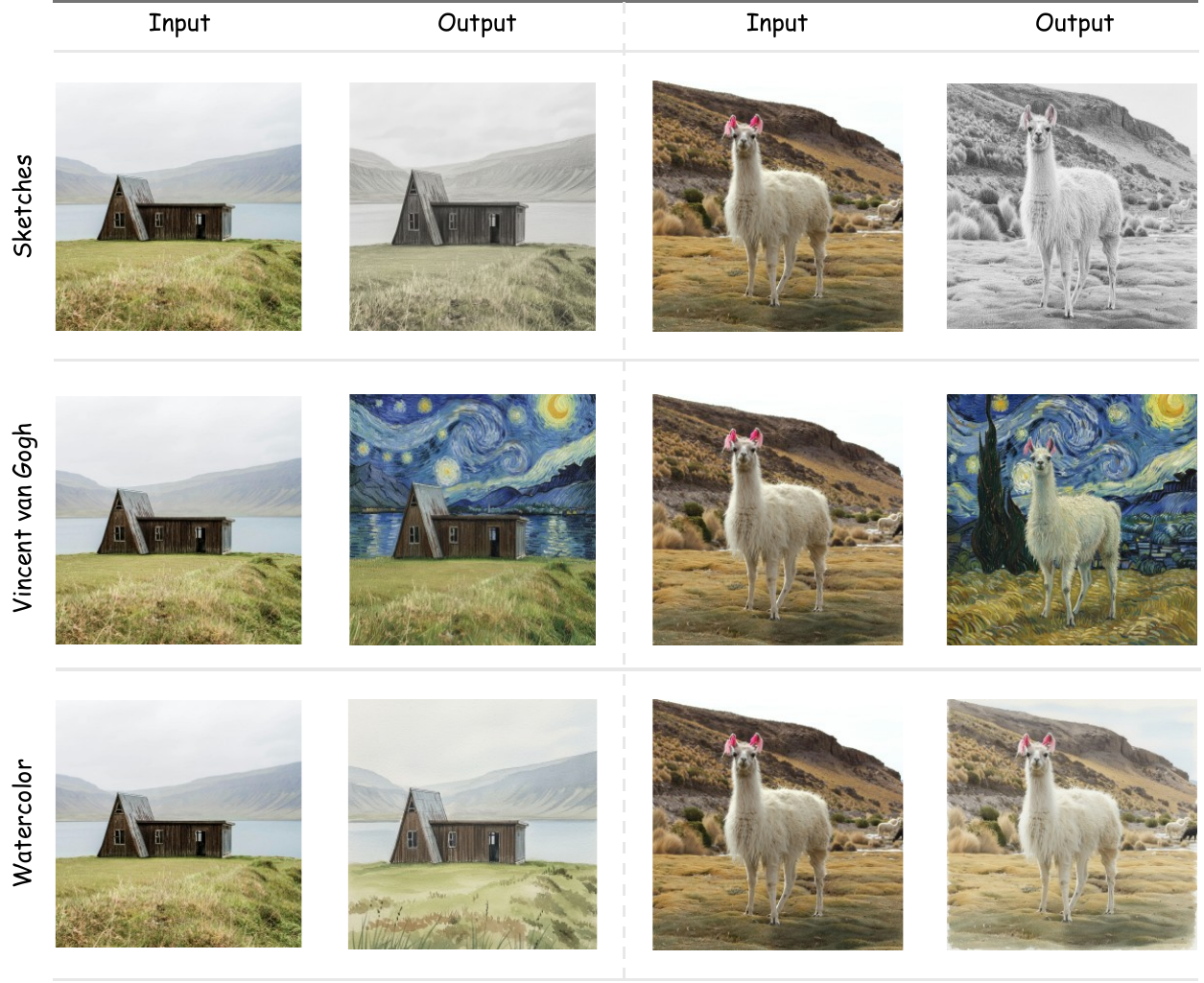}
    \vspace{-1.20em}
    \caption{\textbf{Reference-guided Style Transfer with Structure Preservation.}
  A key advantage of CARE-Edit is the ability to decouple style from structure via expert routing. Unlike holistic transfer methods that often deform the underlying geometry, our approach injects the target aesthetic while strictly anchoring the scene layout. \textbf{(i) Top:} The results show that CARE-Edit's \emph{Reference Expert} successfully translates the scene into the swirling impasto of Van Gogh or a watercolor wash, yet the \emph{Base Expert} ensures the architectural rigidity of the cabin, preserving the straight lines of the roof and window frames.
    \textbf{(ii) Bottom:} The subject's fur texture is re-rendered to match the artistic medium, demonstrating CARE-Edit's successful fine-grained textural adaptation without distorting the animal's original silhouette or pose.}
    \label{fig:A5}
\end{figure*}

In this concluding subsection, we present extensive visual evidence of CARE-Edit across different editing tasks and summarize how they map to practical usage scenarios. Figures~\ref{fig:A3}--\ref{fig:A5} show extended qualitative results for object removal, object addition and replacement, and style transfer.

\vspace{-0.80em}
\paragraph{Object Removal.} 
Given an input image, the model removes the selected object and synthesizes background content consistent with the surrounding regions. Figure~\ref{fig:A3} shows that our CARE-Edit can inpaint relatively large masked areas without obvious seams or blur, while leaving unedited regions nearly unchanged.

\vspace{-0.80em}
\paragraph{Addition and Replacement.} 
The user provides a short text instruction (\eg, “Add …”, “Replace …”) and a coarse mask indicating desired location and approximate size of the edited object. Figure~\ref{fig:A4} shows that CARE-Edit uses this mask to control scale and placement, filling the region with an object that matches the text and blends with the scene.

\vspace{-0.80em}
\paragraph{Style Transfer.} 
The image content is largely preserved, while global appearance is modified according to a target style. Figure~\ref{fig:A5} shows that CARE-Edit maintains scene structure and object boundaries, avoiding severe detail loss.

These results show that CARE-Edit can handle removal, addition, replacement, and stylistic changes.

\section{Analysis of Expert Lattent Attention Maps}
\label{app:analysis}

To complement the qualitative results and provide a mechanistic understanding of CARE-Edit, we conduct a deep diagnostic empirical analysis of the model's internal expert learning behavior. A core hypothesis of this work is that different editing conditions (\eg, text semantics vs. spatial masks vs. reference style) impose different learning dynamics on a shared backbone. CARE-Edit resolves this via an explicit routing of heterogenous, specialized experts.

\begin{figure}[t]
\centering
\includegraphics[width=\linewidth]{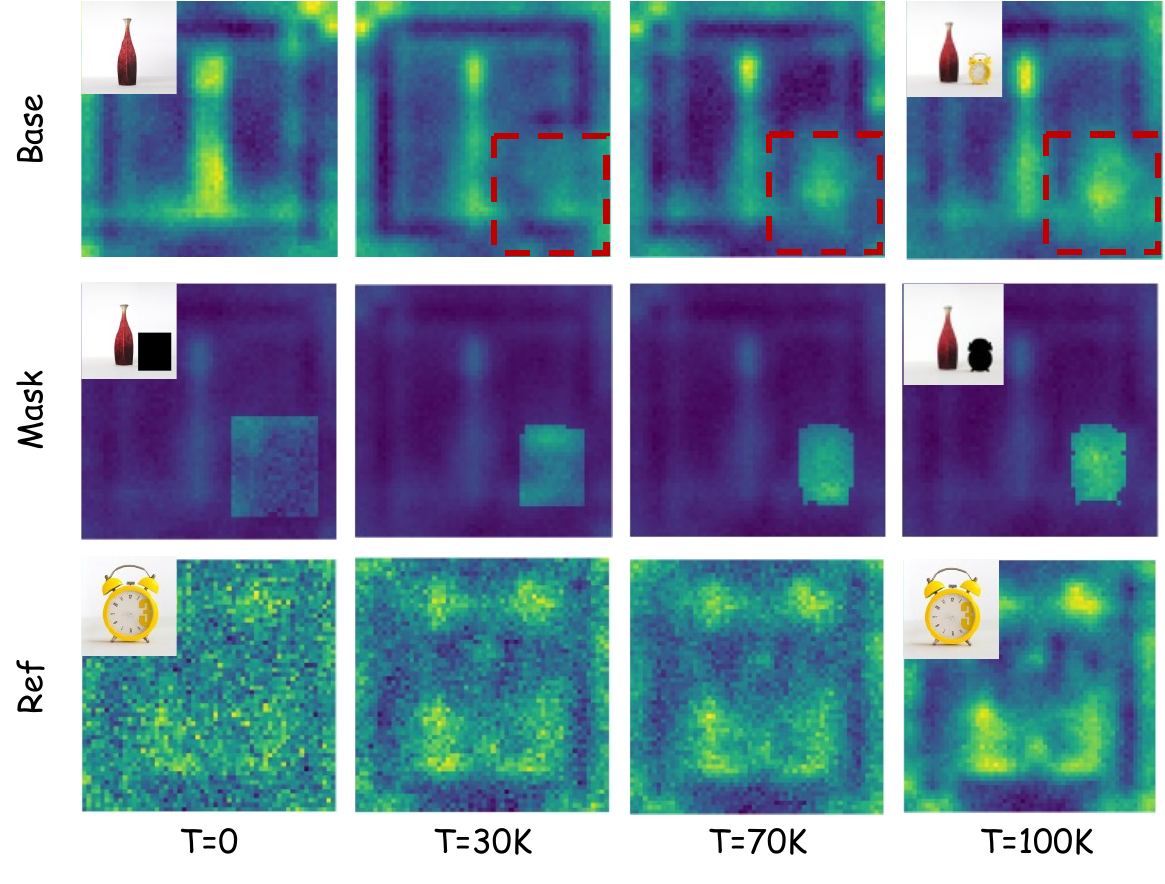}
\vspace{-0.60em}
\caption{\textbf{Expert Specialization During Training.} We visualize the latent attention maps for the \textit{Base}, \textit{Mask}, and \textit{Reference} experts at increasing training iterations ($T$).
\textbf{(Top) Base Expert:} Maintains consistent, global activation throughout training ($T=0 \rightarrow 100\mathrm{K}$), acting as a foundation to preserve image structure.
\textbf{(Middle) Mask Expert:} Learns to progressively suppress background signals, evolving from a noisy initialization to a highly localized, structure-aware attention map that precisely targets the edit region.
\textbf{(Bottom) Reference Expert:} Gradually increases engagement in regions requiring semantic or stylistic modification.
This distinct separation of concerns confirms that CARE-Edit effectively disentangles conflicting editing signals into specialized processing pathways.}
\label{fig:expert_full}
\end{figure}

The main paper only visualizes the \textit{Base} Expert due to the space limitations. To validate that these experts indeed evolve distinct, complementary roles rather than collapsing into a uniform average, we visualize the attention maps of all three condition-aware experts, (i) \textit{Base}, (ii) \textit{Mask}, and (iii) \textit{Reference}, throughout the training process. Figure~\ref{fig:expert_full} illustrates the evolution of these attention map distributions at different training iterations ($T=0$, $T=30\mathrm{K}$, $T=70\mathrm{K}$, and $T=100\mathrm{K}$). This visualization effectively opens the black box of these experts, revealing how the model learns to disentangle complex editing objectives over time.

\paragraph{Base Expert: The Global Anchor.}
As shown in Figure~\ref{fig:expert_full}, the \textit{Base} Expert (Top Row) maintains a robust, spatially widespread activation pattern across the entire training trajectory. Even at late training stages ($T=100\mathrm{K}$), its attention map covers the majority of the image canvas with high intensity. This confirms its role as the task-agnostic anchor, which is responsible for preserving the intrinsic structure, lighting, and layout of the original image, while incorporating conditional information. By handling the global coherence, the \textit{Base} Expert frees the other experts to focus purely on differential changes, ensuring that the unedited regions remain perceptually and semantically consistent with base images.

\paragraph{Mask Expert: Spatial and Geometric Specialization.}
The \textit{Mask} Expert (mid-row in Figure~\ref{fig:expert_full}) displays the most dramatic evolution, demonstrating the emergence of spatial intelligence. At early training stages ($T=0$), the attention is diffuse, noisy, and object-unaware. However, as training progresses through the mid-phase ($T=30\mathrm{K}$ to $70\mathrm{K}$), the entire attention becomes aggressively focused, concentrating strictly within and immediately around the user-provided input editing regions. By $T=100\mathrm{K}$, the \textit{Mask} Expert exhibits fine-grained, binary-like activation boundaries that align perfectly with the intended edit objects. This trajectory indicates that the \textit{Mask} Expert successfully learns to exploit the \textit{Mask Repaint} module's signals, delegating geometric restructuring (\eg, object removal, shape modification) exclusively to this expert while suppressing its influence on the original background to prevent potential leakage or artifacts.

\paragraph{Reference Expert: Semantic and Stylistic Injection.}
The \textit{Reference} Expert (Bottom Row in Figure~\ref{fig:expert_full}) exhibits a distinct pattern of "semantic sparsity." Unlike the Mask expert, which aligns with geometry, the Reference expert aligns with content relevant to the style or identity transfer. Initially inactive, its attention grows as the model learns to map features from the reference image encoder ($Z_r$) to the generated latent space. At convergence stage ($T=100\mathrm{K}$), we observe heightened activation in regions that require texture synthesis or photometric adjustment (\eg, the surface of an object changing material, or the entire scene during style transfer). More importantly, its activation map is orthogonal to the Base expert. It injects fine-grained appearance cues (\eg, color, texture) without overwriting the original structural geometry maintained by the Base and Mask experts.

Overall, these distinct activation signatures observed in Figure~\ref{fig:expert_full} validate the efficacy of our \textit{Condition-Aware Routing} design. Instead of forcing a single set of weights to compromise between preserving identity and changing style, CARE-Edit dynamically distributes the workload: the \textit{Mask} expert handles the ``where," the \textit{Reference} expert handles the ``what" (appearance), and the \textit{Base} expert ensures global consistency of the image to be edited. This learned specialization serves as the key factor enabling our CARE-Edit to minimize task interference from mulitple inputs and thus achieve high-fidelity editing in the challening multi-condition scenarios.

\end{document}